\pdfoutput=1
\documentclass[dvipsnames]{article}
\usepackage{iclr2027_conference,times}

\usepackage{amsmath,amsfonts,bm}

\def\eqref#1{(\ref{#1})}

\def\1{\bm{1}}

\DeclareMathAlphabet{\mathsfit}{\encodingdefault}{\sfdefault}{m}{sl}
\SetMathAlphabet{\mathsfit}{bold}{\encodingdefault}{\sfdefault}{bx}{n}

\usepackage{hyperref}
\usepackage{url}

\usepackage{graphicx}
\usepackage{balance}
\usepackage{amsmath}
\usepackage{algpseudocode}
\usepackage{latexsym}
\usepackage{multirow}
\usepackage{multicol}
\usepackage{color}
\usepackage{dashrule}
\usepackage{extarrows}
\usepackage{dsfont}
\usepackage{centernot}
\usepackage{hyperref}
\usepackage{natbib}
\usepackage{fancybox}
\usepackage[utf8]{inputenc}
\usepackage[T1]{fontenc}
\usepackage{hyperref}
\usepackage{url}
\usepackage{booktabs}
\usepackage{amsfonts}
\usepackage{amsmath}
\usepackage{nicefrac}
\usepackage{microtype}
\usepackage{xcolor}
\usepackage{amsmath}
\usepackage{lipsum}
\usepackage{colortbl}
\usepackage{subcaption}
\usepackage{tabularx}
\usepackage{makecell}
\usepackage{wrapfig}
\usepackage{bm}
\usepackage{multirow}
\usepackage{xcolor}
\usepackage{bbm}
\usepackage{mdframed}
\usepackage{enumitem}
\usepackage{xspace}

\usepackage{lineno}

\definecolor{darkblue}{HTML}{991B1B}
\hypersetup{colorlinks=true, citecolor=darkblue, linkcolor=darkblue, urlcolor=darkblue}

\usepackage{listings}
\usepackage{xcolor}
\usepackage[ruled,vlined]{algorithm2e}
\usepackage{caption}
\usepackage{float}
\usepackage{wrapfig}
\usepackage{tcolorbox}
\tcbuselibrary{theorems}
\tcbuselibrary{most}
\usepackage{xcolor}
\usepackage{hyperref}
\usepackage{url}
\usepackage{booktabs}
\usepackage{amsfonts}
\usepackage{nicefrac}
\usepackage{microtype}
\usepackage{xcolor}
\usepackage{amsmath}
\usepackage{multirow}
\usepackage{multicol}
\usepackage{colortbl}
\usepackage{tcolorbox}
\usepackage{wrapfig}
\usepackage{cleveref}
\usepackage{xspace}
\usepackage{graphicx}
\usepackage{url}
\usepackage{booktabs}
\usepackage{hyperref}
\usepackage{graphicx}
\usepackage{microtype}
\usepackage{multicol}
\usepackage{xcolor}
\usepackage{multirow}
\usepackage{lipsum}
\usepackage{enumitem}
\usepackage{pifont}
\usepackage{colortbl}
\usepackage{tabularx}
\usepackage{caption}
\usepackage{float}
\usepackage{amsmath,amsfonts,amssymb,bbm}
\usepackage{adjustbox}
\usepackage{tikz}
\usetikzlibrary{arrows.meta, positioning, calc, fit, backgrounds}

\usepackage{lineno}

\definecolor{darkblue}{HTML}{991B1B}
\hypersetup{colorlinks=true, citecolor=darkblue, linkcolor=darkblue, urlcolor=darkblue}
\hypersetup{pdftitle={Tev: Scaling Long-Horizon Agentic Environments with Dense Progress Rewards},
  pdfauthor={Wenbo Hu, Shraman Pramanick, Krishna Kumar Singh, Nanyun Peng, Kai-Wei Chang, Yong Jae Lee}}
\definecolor{light-gray}{gray}{0.6}
\definecolor{front-color}{HTML}{F5FFFA}
\definecolor{Gray}{gray}{0.93}
\definecolor{customTeal}{RGB}{0, 128, 128}
\definecolor{emphasisColor}{RGB}{255, 0, 0}
\definecolor{oursBlue}{RGB}{51,202,246}
\definecolor{deltaBlue}{HTML}{C8102E}

\definecolor{blue1}{HTML}{508AB2}
\definecolor{green2}{HTML}{BFF6BA}
\definecolor{closed_models}{RGB}{255, 219, 187}
\definecolor{open_models_below_4B}{RGB}{185, 235, 255}
\definecolor{open_models_7B_12B}{RGB}{39, 121, 1}

\definecolor{columngreen}{HTML}{D4F2D7}
\definecolor{columnblue}{HTML}{C2E6F5}
\definecolor{columnred}{HTML}{F5C2CC}
\definecolor{columnpurple}{HTML}{E6D4F2}
\definecolor{columnyellow}{HTML}{FBE0BC}
\definecolor{columncyan}{HTML}{C2F5E6}
\definecolor{columnpeach}{HTML}{F5D4C2}
\definecolor{oursrow}{HTML}{FEE2E2}
\newcommand{\model}{\textsc{Tev}\xspace}
\newcommand{\dataset}{\textsc{Tev-Gym}\xspace}
\newcommand{\evalsuite}{\textsc{ALE-CM}\xspace}

\title{\model: Scaling Long-Horizon Agentic\\ Environments with Dense Progress Rewards}

\newcommand{\demourl}{https://tev-agent.github.io/}
\newcommand{\linkicon}[1]{\raisebox{-0.18em}{\includegraphics[height=1.05em]{#1}}}

\author{Wenbo Hu$^{1,2}$\quad Shraman Pramanick$^{1}$\quad Krishna Kumar Singh$^{1}$ \\
\textbf{Nanyun Peng$^{2}$\quad Kai-Wei Chang$^{2}$\quad Yong Jae Lee$^{1}$} \\
$^{1}$Adobe Research\quad $^{2}$University of California, Los Angeles \\[4pt]
\linkicon{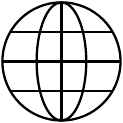}\,{\bfseries\href{\demourl}{\nolinkurl{https://tev-agent.github.io}}}
}

\iclrfinalcopy
\begin{document}
\addtocontents{toc}{\protect\setcounter{tocdepth}{-1}}

\maketitle
\lhead{Preprint}

\begin{abstract}

Long-horizon agents have drawn growing attention for their ability to work autonomously---from hours to days---on economically valuable tasks, a capability essential to automated discovery and recursive self-improvement (RSI). Yet open-source models still struggle on long-horizon tasks, largely for lack of both training data and a working recipe, and scaling both is hard: long, complex tasks are difficult to curate and verify, and their rewards are sparse. We address both. First, we introduce \dataset, the \textit{first} long-horizon agent training dataset with verifiable, dense, progressive rewards---$1{,}170$ verified task environments across $7$ domains.
It is built on a \emph{latent-world-first} approach that first constructs a \emph{latent world}: a complete and correct hidden specification of the task domain, workflow, and expected state after each step. Each task is then progressively rendered from this specification into an agent-facing environment, enabling controllable complexity and a natural progress metric.
Second, we turn this metric into a training recipe: via progress-potential reward shaping (PPRS), we convert progress into a dense per-step reward that improves training efficiency and makes otherwise-intractable tasks trainable. Training with this recipe yields \model-27B, which achieves substantial gains on the Agents' Last Exam computing and math sciences domains (ALE-CM), matching GPT-6 Astra and Claude Opus 5 at Pass@5. Despite training only on ALE-related data, it generalizes to the Terminal-Bench (TB) 3.0 and 4.0 subsets, surpassing GLM 5.2 and Gemini 3.7 Flash, and reaches Claude Opus 4.8-level performance on middle-horizon tasks such as TB 2.1.

\end{abstract}
\begin{figure}[H]
\vspace{-5mm}
\centering
\includegraphics[width=0.97\textwidth]{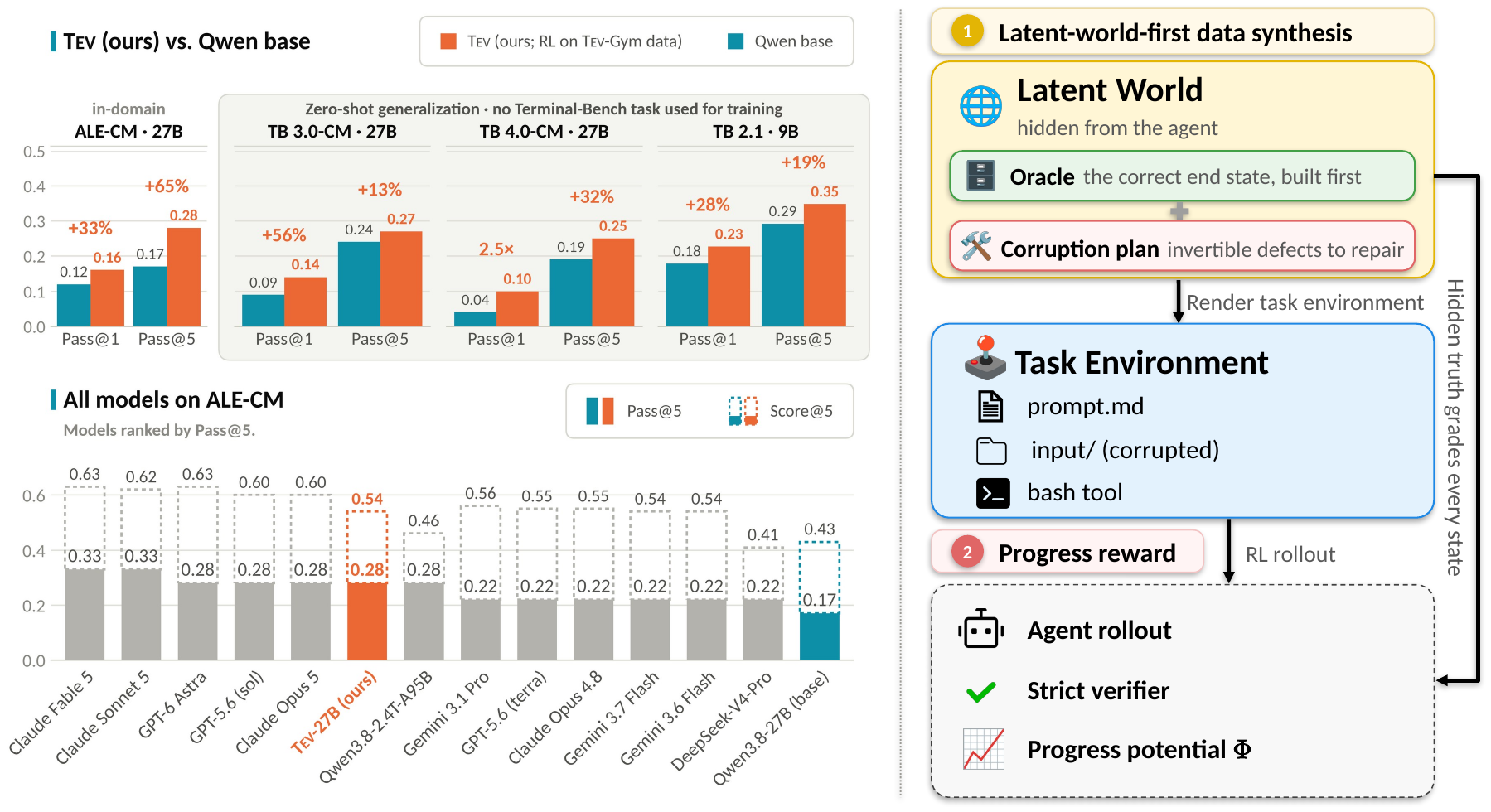}
\vspace{-2mm}
\caption{We demonstrate \model's results on ALE-CM, and generalization to Terminal-Bench (TB) 3.0-CM, 4.0-CM, and 2.1 on the left and a summary of our \textit{latent-world-first} data synthesis approach on the right. \model showcases significant performance gains across two model scales and \model-27B matches GPT-6 Astra and Claude Opus 5 at Pass@5 on ALE-CM.}
\label{fig:teaser}
\end{figure}

\section{Introduction}
\label{sec:intro}
\vspace{-2mm}

Recent progress in large language models (LLMs) has shifted attention toward long-horizon
agents~\citep{openai2026gpt6,anthropic2026fable,moonshot2026kimik3}, which can work autonomously for hours to days on
complex, economically valuable tasks~\citep{sun2026agentslastexam,styles2024workbench,zhu2026edgebench}.
Long-horizon tasks, such as solving open problems in mathematics and science, conducting automated machine learning research, and advancing coding and security systems, are fundamental to automated discovery and recursive self-improvement (RSI).

\begin{figure}[t]
\vspace{-8mm}
\centering
\includegraphics[width=0.92\textwidth]{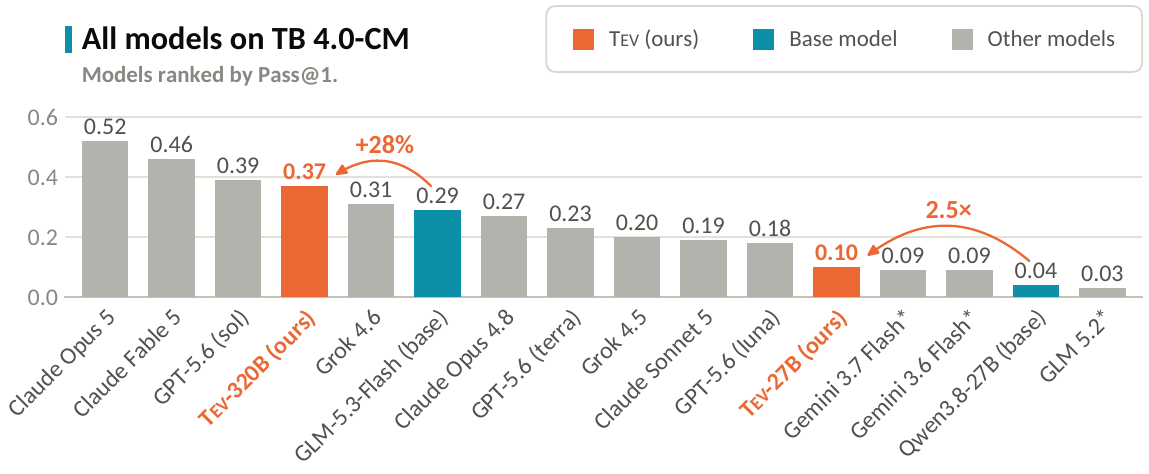}
\vspace{-1mm}
\caption{\textbf{All models on the Terminal-Bench 4.0 ``CM'' subset (TB 4.0-CM), ranked by Pass@1.} \model-320B and \model-27B are trained with our recipe from GLM-5.3-Flash and Qwen3.8-27B, respectively; each arrow gives the relative gain over its base.}
\vspace{-3mm}
\label{fig:tb4-ranking}
\end{figure}

Despite these promising breakthroughs, coding-agent progress in the open-source community has focused on shorter- and
mid-horizon tasks, scaled via large agentic training corpora such as SWE-Smith~\citep{yang2025swesmith},
TMAX~\citep{ivison2026tmax}, Endless Terminals~\citep{gandhi2026endless}, CLI-Gym~\citep{lin2026cligym},
SETA~\citep{shen2026seta}, and Terminal-Universe~\citep{wu2026terminaluniverse}. These datasets contain thousands to tens of thousands of environments, but each involves only a pull-request-sized fix or a short terminal session, and their unit-test-based evaluation provides only a terminal pass/fail signal or the fraction of tests passed. By contrast, genuinely long-horizon tasks are available only in human-authored benchmarks~\citep{merrill2026terminalbench,li2026lhterminalbench}. These benchmarks contain only tens to a few hundred instances, each requiring weeks of expert effort. They provide partial credit only through hand-written checkpoints and exist only as evaluation sets. Thus, no training corpus of long-horizon tasks exists: agentic training corpora are large but short-horizon, while long-horizon tasks exist only as small, human-authored evaluation sets.

Curating long-horizon tasks requires balancing two competing goals: tasks must be sufficiently long and structurally complex to provide meaningful training while remaining exactly verifiable~\citep{zeng2026ace}. Scaling long-horizon environments for training is even more difficult because rewards are sparse and available only upon task completion, depriving long-tail tasks of useful training signals. To address both limitations, we introduce a \emph{latent-world-first} synthesis approach that begins by constructing a \emph{latent world}, a complete and correct hidden specification of the task domain.
We first represent the underlying workflow as a directed acyclic graph (DAG) within a latent world and then progressively render the input files from that world. This procedure allows us to control task complexity while retaining every correct intermediate state, which provides a natural metric for measuring progress and therefore a verifiable dense reward for our training. Unlike pipelines that ask a language model to generate the task, inputs, and checker without a true answer, our latent world serves as an exact answer key for reliable verification.

Long-horizon tasks come with sparse rewards, arriving only at task completion and submission, which notoriously starves
training on long-tail samples~\citep{yu2025dapo,shen2026seta,zeng2026ace}. Group-relative policy optimization~\citep[GRPO;][]{shao2024deepseekmath}
normalizes rewards within a group of $G$ rollouts of the same task, so under terminal-only reward, a
group in which all $G$ rollouts fail has zero advantage everywhere and contributes nothing to the
gradient. Mid-horizon pipelines can afford to discard such groups~\citep{ivison2026tmax,yu2025dapo}
because they are rare there; on long-horizon tasks they are most common at the start of training,
when the policy is weakest and signal is needed most. We therefore introduce progress-potential reward
shaping (PPRS) into reinforcement learning (RL), which makes training efficient and makes otherwise-intractable tasks trainable. PPRS scores the agent's workspace after every turn $t$ against the
known-correct intermediate states from our latent world, giving a progress potential $\Phi(s_t)\in[0,1]$ that rises as the
workflow is completed and falls on regressions.
Reward shaped by $\Phi$ gives within-group variance from the first turn, even
when no rollout finishes.

Figure~\ref{fig:teaser} summarizes our approach and \model's results, and Figure~\ref{fig:tb4-ranking} ranks all models on the Terminal-Bench (TB)~4.0 ``CM'' subset by Pass@1.
Using our latent-world-first synthesis approach, we introduce \dataset, composed of 1{,}170
verified long-horizon task environments across 7 domains (software engineering, data engineering,
AI/CS research, mathematics and operations research, security and forensics, infrastructure
administration, and quantum computing). As illustrated in
Table~\ref{tab:data-comparison}, \dataset's tasks take a frontier agent a median of 12.4 minutes per rollout, about 6 times longer than the most time-intensive prior corpus~\citep{yang2025swesmith}, and have the lowest pass rate, demonstrating our training data's complexity and quality. Moreover, \dataset can be scaled beyond a fixed collection by coding agents writing generators that produce new verified environments across seeds and scale tiers without further model calls.

\begin{table}[t]
\vspace{-8mm}

\centering
\small
\caption{\textbf{\dataset versus prior agentic task corpora.} ``Agent time'' is the median and
mean wall-clock minutes per rollout of Gemini~3.6~Flash, from first action to last step
(details in Appendix~\ref{app:data:comparison}).
Difficulty is Pass@$k$ measured by Gemini~3.6~Flash across all datasets.
}
\vspace{-3mm}
\label{tab:data-comparison}
\resizebox{\textwidth}{!}{%
\begin{tabular}{l r c rr l l c c c}
\toprule
\multirow{2}{*}{Dataset} & \multirow{2}{*}{Size} & \multirow{2}{*}{\# Domains} & \multicolumn{2}{c}{Agent time (min)} & \multirow{2}{*}{Oracle} & Grading & Progressive & \multirow{2}{*}{P@1} & \multirow{2}{*}{P@8} \\
\cmidrule(lr){4-5}
 & & & Median & Mean & & Metric & Reward & & \\
\midrule
SWE-Smith~\citep{yang2025swesmith}                 & 50.1k & 1 & 2.0     & 2.4    & unit tests   & pass/fail     & \ding{55} & 47\% & 52\% \\
Terminal-Universe~\citep{wu2026terminaluniverse}   & 32.0k & 3 & NA      & NA     & pytest       & pass/fail     & \ding{55} & NA  & NA \\
TMAX-15K~\citep{ivison2026tmax}                    & 14.6k & 3 & 1.3     & 2.1    & unit tests   & pass/fail     & \ding{55} & 74\% & 80\% \\
SETA~\citep{shen2026seta}                          & 4.5k  & 4 & 1.2     & 2.1    & unit tests   & test fraction & \ding{55} & 80\% & 88\% \\
Endless Terminals~\citep{gandhi2026endless}        & 3.3k  & 2 & 0.5     & 0.7    & unit tests   & pass/fail     & \ding{55} & 83\% & 89\% \\
CLI-Gym~\citep{lin2026cligym}                      & 1.7k  & 4 & 1.9     & 2.3    & unit tests   & pass/fail     & \ding{55} & 69\% & 74\% \\
\midrule
\rowcolor{oursrow}
\textbf{\dataset (ours)} & 1{,}170 & \textbf{7} & \textbf{12.4} & \textbf{14.8} & latent world & staged $\Phi$ $+$ pass/fail & \ding{51} & \textbf{29\%} & \textbf{40\%} \\
\bottomrule
\end{tabular}%
}
\vspace{-3mm}
\end{table}

We introduce \model-9B and \model-27B, trained with PPRS on the unit tier and the full corpus of \dataset, respectively. On \evalsuite, the computing-and-math split of Agents' Last Exam~\citep{sun2026agentslastexam}, \model-9B improves Pass@5 by $83\%$ over Qwen3.5-9B, while \model-27B improves it by $65\%$ over Qwen3.8-27B to $0.28$, matching GPT-6 Astra and Claude Opus 5. Despite training only on ALE-related data, \model-27B generalizes to the TB~3.0 and 4.0 subsets~\citep{merrill2026terminalbench,marten2026tb3,marten2026tb4}, improving Pass@1 by $56\%$ and $150\%$, respectively, and outperforming GLM~5.2 and Gemini~3.7 Flash. On the mid-horizon TB~2.1, \model-9B improves zero-shot Pass@1 by $28\%$, while \model-27B reaches $0.794$ Pass@1, matching Claude Opus~4.8-level performance. The recipe also transfers to a larger base: applying it to GLM-5.3-Flash yields \model-320B, which raises Pass@1 on the TB~3.0 and 4.0 subsets by $35\%$ and $28\%$ and, on TB~4.0, trails only Claude Opus~5, Claude Fable~5, and GPT-5.6 (sol).

Our main contributions are summarized as follows:
\vspace{-1mm}
\begin{itemize}[align=right,itemindent=0em,labelsep=2pt,labelwidth=1em,leftmargin=*,itemsep=0em]
\item We propose \textbf{latent-world-first synthesis}, a pipeline that constructs long-horizon agentic tasks backward from a known-correct world, giving exact verification, a
controllable difficulty ladder, and a progressive reward in the middle.
\item We introduce \textbf{\dataset}, the \textit{first} long-horizon agent training dataset with progressive rewards, composed of
1{,}170 verified long-horizon task environments across 7 domains.
\item We present \textbf{\model-27B}, trained on \dataset using PPRS, and demonstrate its superior performance on \evalsuite, the TB~3.0 and TB~4.0 subsets, and TB~2.1; the same recipe applied to GLM-5.3-Flash gives \textbf{\model-320B}, which improves further.
\end{itemize}
\vspace{-1mm}

\section{Long-Horizon Environment Synthesis}
\label{sec:data}
\vspace{-2mm}

\begin{figure}[t]
\vspace{-8mm}

\centering
\includegraphics[trim=0cm 6.1cm 2.36cm 0cm, clip, width=\linewidth]
{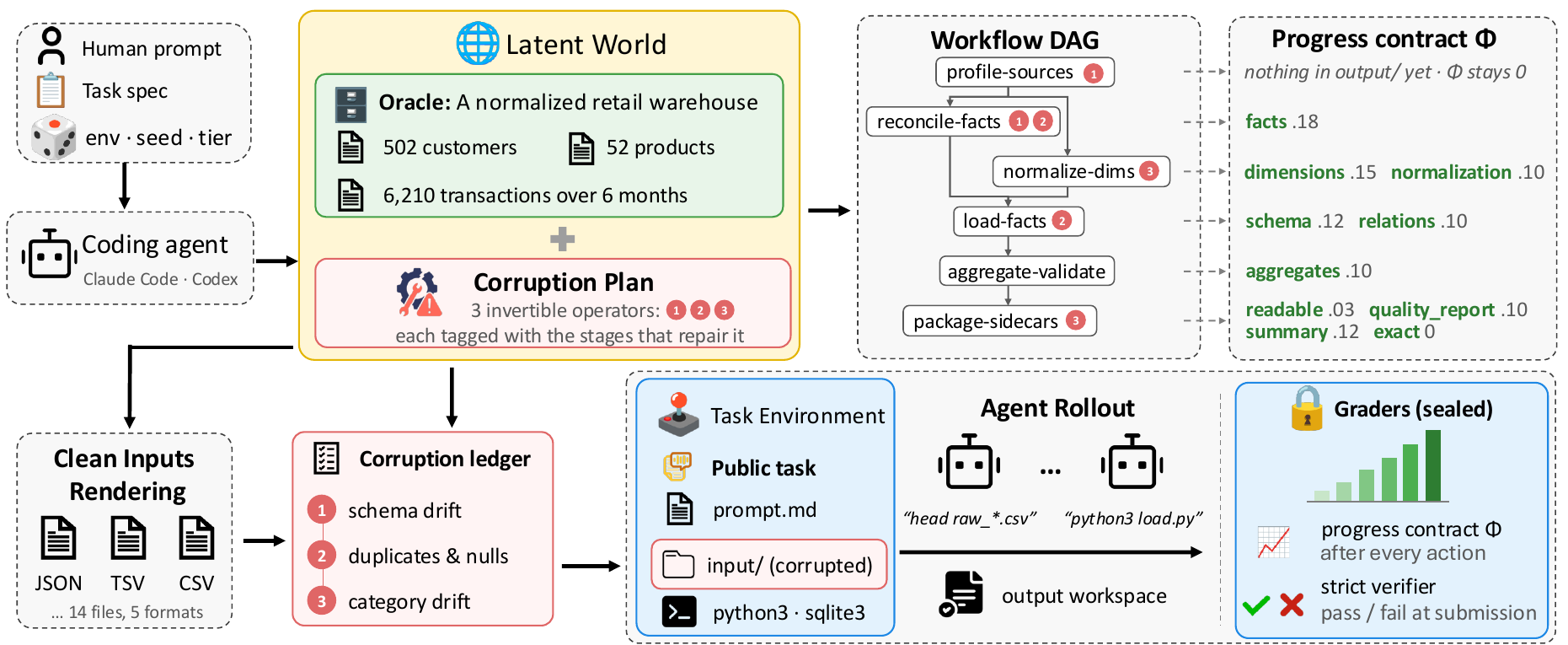}
\vspace{-4mm}
\caption{\textbf{Latent-world-first synthesis, on one example \dataset environment.} This environment's task is ``Retail Warehouse ETL Recovery. Goal: build a clean star-schema SQLite warehouse from six months of messy retail data and produce two truthful JSON sidecars.''
}
\vspace{-3mm}
\label{fig:data-pipeline}
\end{figure}

Figure~\ref{fig:data-pipeline} details the mechanisms of long-horizon environment reconstruction.
This section defines how an environment is synthesized backward from a latent world (\S\ref{sec:data:pipeline}), and how coverage, difficulty, and
correctness are enforced in \dataset (\S\ref{sec:data:quality}).

\subsection{Latent-World-First Task Synthesis}
\vspace{-2mm}
\label{sec:data:pipeline}

\textbf{Stage 0: task sources.} Our training data targets \evalsuite, whose coverage we group into
seven major domains. Within each domain, we randomly sample specifications and author our own
generator around each one. For details, please refer to Appendix~\ref{app:data:sources}.

\textbf{Stage 1: world construction.} Each generator is a seeded program. Given a seed and a scale
tier, it procedurally generates the latent world $W$, the complete, correct state of the domain
(e.g., a retail warehouse's customers, products, and transactions, or a distributed checkpoint's
tensor shards and topology), and we validate $W$ against the environment's invariants (unique
identities, valid joins, finite measures). The same step draws a seed-derived \emph{defect plan}:
exactly which records, shards, or fields of $W$ will later be corrupted, and by which declared
operator.

\textbf{Stage 2: workflow graph.} We compile the environment's workflow into a directed acyclic
graph (DAG) of stages with explicit dependencies, permitted parallelism, and per-stage
semantic-work estimates (e.g., \texttt{profile-sources} $\rightarrow$ \texttt{reconcile-facts} /
\texttt{normalize-dimensions} $\rightarrow$ \texttt{load-facts} $\rightarrow$
\texttt{aggregate-validate} $\rightarrow$ \texttt{package-sidecars}). The DAG is built by code
from the scale of $W$, with no model in the loop. Its critical path, the longest chain of
dependent stages, defines the task's horizon, and its stages form the skeleton of the progress
potential in \S\ref{sec:dense}.

\textbf{Stage 3: artifact rendering.} We \emph{project} $W$ onto the environment's declared public
interface, materializing the clean input files in their target wire formats: CSV and TSV
partitions, JSON records, SQLite databases, \texttt{safetensors} shards, stripped binaries, or
infrastructure manifests. Rendering is a pure function of $W$, and before proceeding we assert
that it round-trips, i.e., that $W$ is recovered exactly from the rendered files. Together with the invertible corruption of Stage~4, this makes the agent's
workspace $s_0$ fully determine $W$, so the correct answer is computable from what the agent
actually sees, and the ground truth is correct by construction.

\textbf{Stage 4: controlled corruption.} We apply the Stage-1 defect plan to the rendered files,
injecting schema drift, duplicate and missing records, categorical drift, scrambled shard indices,
stale caches, and the like (Figure~\ref{fig:data-pipeline} shows the three operators of one
environment). The corrupted tree $s_0$ \emph{is} the task the agent receives, and the untouched
world $W$ \emph{is} the ground truth. Because rendering round-trips and corruption is declared and
invertible, $s_0$ fully determines $W$: the correct solution is computable from what the agent
sees and is known exactly without running a solver. More generally, every component of an
environment,
\begin{equation}
E \;=\; \bigl(x,\; s_0,\; \operatorname{ver}_W,\; \Phi_W\bigr),
\label{eq:env}
\end{equation}
is a function of the same $W$: the task prompt $x$, the input tree $s_0$, the strict
pass/fail verifier $\operatorname{ver}_W$, which checks the agent's output files against $W$, and
the progress potential $\Phi_W$ of \S\ref{sec:dense}. This single anchor prevents \emph{artifact
drift}~\citep{ivanov2026anchor}, where an environment, task, and verifier generated separately
each look valid on their own yet describe different problems.

\textbf{Stages 5 and 6: public specification and hidden suite.} We emit the public contract, the
task prompt $x$ (input roles, required output artifacts and their schemas, ordered requirements,
authority-precedence rules, and the permitted toolchain), which is the only description the agent
sees. Outside the agent's view, we seal a hidden suite used at verification time: metamorphic
transforms, which are semantics-preserving rewrites that must not change the grade, and
stale-output counterfactuals, which are plausible-but-wrong states that must fail.

\textbf{Generalization of latent-world-first synthesis.} We focus on the \evalsuite domains due to resource constraints, but the approach applies to other ALE domains. Appendix~\ref{app:data:examples} walks through three environments built by the pipeline: a multimodal AWS cost-optimization task with its billing-dashboard screenshot rendered from our latent world (cloud operations), closed-loop building control of an EnergyPlus model (engineering), and financial-statement reconstruction from SEC-style filings (finance).

\subsection{Coverage, Calibration, and Qualification}
\vspace{-2mm}
\label{sec:data:quality}

\textbf{Diversity.} Every environment is labeled on 19 coverage axes
(Appendix~\ref{app:data:coverage}), including subdomain, scale, the four horizon axes, workflow
stage, tool capability, artifact role, relation type, and corruption class. The axes that matter
for training are behavioral: changing their value changes what the agent must repair (corruption
class), how (workflow stage, tool capability), and through which files (artifact role, relation
type). For more details, please see our Appendix~\ref{app:data:coverage}.

\textbf{Controllable difficulty.} Two controls change a task's difficulty. The scale
tier (by default \emph{unit}, \emph{small}, or \emph{full}) grows $W$
(Appendix~\ref{app:data:scale}) and lengthens the operational horizon: how much must be read,
joined, and written. The workflow DAG of Stage~2, together with the number of corruption classes
in the defect plan, fixes the dependency depth: how many stages must be completed, and in which
order, before any output can be right. Appendix~\ref{app:data:difficulty} states how we measure
horizon and difficulty and reports base-model and frontier solve rates.

\textbf{Qualification gates and cleanliness.} A synthesized environment is admitted only if it
passes all eight benchmark-independent gates, checked automatically at generation time and each
matched to a named failure class (Table~\ref{tab:data-gates}). Three of them back the progress
potential of \S\ref{sec:dense:potential}: all 14 gaming probes must score $0$ (\emph{exploit
resistance}), the potential must lie strictly between $0$ and $1$ on partially correct states and
increase monotonically to $1$ along the reference trajectory (\emph{progress validity}), and the
progress scorer, strict verifier, and reference solvers must have disjoint code closures
(\emph{role independence}). Appendix~\ref{app:data:gates} states every
gate formally, together with the verifier-independent solver audit, and
Appendix~\ref{app:data:decontamination} describes the structural decontamination check.

\begin{figure}[t]
\vspace{-8mm}

\centering
\definecolor{phiZero}{HTML}{8C8C8C}
\definecolor{phiLow}{HTML}{56B4E9}
\definecolor{phiMid}{HTML}{009E73}
\definecolor{phiHigh}{HTML}{0072B2}
\definecolor{phiVhigh}{HTML}{CC79A7}
\definecolor{phiFull}{HTML}{E69F00}
\definecolor{phiTerm}{HTML}{D62728}
\begin{tikzpicture}[font=\small, x=0.3016cm, y=2.7cm]
  \draw[phiMid, dashed, line width=0.8pt] (0,0.26) -- (31.5,0.26) node[right, font=\scriptsize, align=left, phiMid!85!black] {every table partially loaded\\(progress 1)};
  \draw[phiHigh, dashed, line width=0.8pt] (0,0.69) -- (31.5,0.69) node[right, font=\scriptsize, align=left, phiHigh!85!black] {schema and facts complete\\(progress 2)};
  \draw[phiFull, dashed, line width=0.8pt] (0,1.0) -- (31.5,1.0) node[right, font=\scriptsize, align=left, phiFull!85!black] {all nodes satisfied\\(progress 3)};
  \draw[->] (0,0) -- (31.5,0) node[right] {turn $t$};
  \draw[->] (0,0) -- (0,1.15) node[above] {$\Phi(s_t)$};
  \foreach \y/\l in {0/0, 0.5/0.5, 1/1} {\draw (0,\y) -- (-0.4,\y) node[left] {\l};}
  \foreach \x in {0,5,10,15,20,25,30} {\draw (\x,0) -- (\x,-0.03) node[below, font=\scriptsize] {\x};}
  \draw[line width=1.1pt, phiTerm, dashed] (0,0.02) -- (30,0.02);
  \node[above right, font=\scriptsize, phiTerm, inner sep=1.5pt] at (0.4,0.02) {terminal-only reward: $0$ for all 8 rollouts};
  \node[font=\scriptsize, anchor=west, align=left, inner sep=2pt] at (0.6,0.475) {$\bullet$ workspace complete, never submitted\\$\circ$ submitted, incomplete\\$\times$ context exhausted};
  \draw[line width=1.1pt, phiHigh, opacity=0.85] (-0.43,0.000) -- (14.57,0.000) -- (14.57,0.690) -- (19.57,0.690);
  \draw[phiHigh, line width=1.3pt, fill=white] (19.57,0.69) circle (2.7pt);
  \draw[line width=1.1pt, phiTerm, opacity=0.85] (0.00,0.000) -- (23.00,0.000);
  \node[phiTerm, font=\fontsize{10.5pt}{12pt}\selectfont] at (23.00,0.0) {$\times$};
  \draw[line width=1.5pt, phiFull, opacity=0.85] (0.00,0.000) -- (28.00,0.000) -- (28.00,1.000) -- (28.00,1.000);
  \fill[phiFull] (28.00,1.0) circle (3pt);
  \draw[line width=1.1pt, phiHigh, opacity=0.85] (0.43,0.000) -- (15.43,0.000) -- (15.43,0.690) -- (20.43,0.690);
  \draw[phiHigh, line width=1.3pt, fill=white] (20.43,0.69) circle (2.7pt);
  \draw[line width=1.1pt, phiMid, opacity=0.85] (0.00,0.000) -- (19.00,0.000) -- (19.00,0.260) -- (19.00,0.260);
  \node[phiMid, font=\fontsize{10.5pt}{12pt}\selectfont] at (19.00,0.26) {$\times$};
  \draw[line width=1.1pt, phiHigh, opacity=0.85] (0.00,0.000) -- (21.00,0.000) -- (21.00,0.260) -- (22.00,0.260) -- (22.00,0.690) -- (30.00,0.690);
  \node[phiHigh, font=\fontsize{10.5pt}{12pt}\selectfont] at (30.00,0.69) {$\times$};
  \draw[line width=1.1pt, phiTerm, opacity=0.85] (0.00,0.000) -- (29.00,0.000);
  \node[phiTerm, font=\fontsize{10.5pt}{12pt}\selectfont] at (29.00,0.0) {$\times$};
  \draw[line width=1.1pt, phiTerm, opacity=0.85] (0.00,0.000) -- (28.00,0.000);
  \node[phiTerm, font=\fontsize{10.5pt}{12pt}\selectfont] at (28.00,0.0) {$\times$};
\end{tikzpicture}
\vspace{-3mm}
\caption{\textbf{The progress potential along one group ($G{=}8$) of rollouts.} Eight rollouts of the trained
27B model on one warehouse environment, graded live after every turn. Under terminal-only reward (red) every rollout scores $0$. Under progressive rewards, by contrast, two submit an
incomplete warehouse (circles), five exhaust the 65k-token context (crosses), and one builds a
complete warehouse, so the rewards vary within the group.
}
\label{fig:potential}
\vspace{-3mm}
\end{figure}
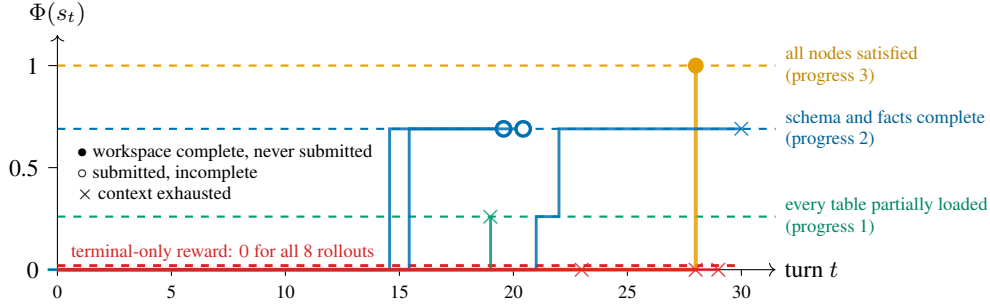

\section{Dense Reward for Long-Horizon RL}
\label{sec:dense}
\vspace{-2mm}

The latent world that grades task completion also grades its middle. This section turns it into a training signal: we quantify why terminal-only reward starves group-relative RL (\S\ref{sec:dense:sparsity}), define the progress potential every \dataset environment provides (\S\ref{sec:dense:potential}), and detail the shaped reward (\S\ref{sec:dense:shaping}).

\subsection{Terminal Reward Starves Group-Relative RL}
\vspace{-2mm}
\label{sec:dense:sparsity}

Consider GRPO~\citep{shao2024deepseekmath}, the estimator we and most recent agentic RL pipelines
use. For a task $x$ the policy samples $G$ rollouts $\tau_1,\dots,\tau_G$ with rewards
$R_1,\dots,R_G$, and each rollout is weighted by its group-normalized advantage
\begin{equation}
\hat A_i \;=\; \frac{R_i-\operatorname{mean}(R_{1:G})}{\operatorname{std}(R_{1:G})+\epsilon}.
\label{eq:grpo-adv}
\end{equation}
With terminal reward $R_i=\mathds{1}[\tau_i\ \text{passes}]$, where a rollout \emph{passes} when the
policy issues the submit action and the strict verifier $\operatorname{ver}_W$ accepts the submitted
workspace, a group in which every rollout
fails has $\hat A_i=0$ for all $i$: the task contributes nothing to the gradient and its $G$
rollouts are wasted. If the policy solves the task with probability $p$, the fraction of wasted groups is $(1-p)^G$: $66\%$ at $p{=}0.05$ and $G{=}8$, and $92\%$ at
$p{=}0.01$ (Appendix~\ref{app:dense}). On long-horizon tasks, where $p$ for an untrained open model
is roughly one in ten or lower (\S\ref{sec:experiments}), dropping or resampling such
groups~\citep{ivison2026tmax,yu2025dapo} discards most of the batch. A group with a single success
fares only slightly better. Its one positive rollout is contrasted with $G-1$ failures that all
receive the same negative advantage, so the gradient cannot tell a near-miss from a rollout that
made no progress.

\subsection{A Progress Potential from the Latent World}
\vspace{-2mm}
\label{sec:dense:potential}

Every \dataset environment provides a \emph{progress contract}: a set of nodes $V$, each a checkable
state of the output workspace derived from the workflow DAG of \S\ref{sec:data:pipeline}, with
weights $w_v>0$, $\sum_v w_v=1$, and prerequisite sets $\mathrm{pre}(v)$ (example in
Appendix~\ref{app:dense}). A \emph{partial scorer} $c_v(s)\in[0,1]$ grades
the agent's current workspace $s$ on node $v$ directly against the latent world. Credit is prerequisite-gated: $\tilde c_v(s)$ is the lowest
partial score among $v$ and all its direct and indirect prerequisites. The potential, written $\Phi$
for $\Phi_W$ of Eq.~\eqref{eq:env}, is the weighted sum:
\begin{equation}
\Phi(s)\;=\;\sum_{v\in V} w_v\,\tilde c_v(s)\;\in\;[0,1].
\label{eq:potential}
\end{equation}

\textbf{Preventing reward hacking.} Three properties make $\Phi$ a usable training signal, each enforced by a qualification gate
(\S\ref{sec:data:quality}): it is \emph{anchored} (exploit resistance), with
the untouched public tree and every gaming probe scoring $\Phi=0$ and $\Phi=1$ exactly when the
strict verifier accepts the workspace; \emph{monotone} (progress validity), rising as nodes are completed and
falling when completed work is destroyed; and \emph{independent} (role independence), computed by a
scorer whose code closure is disjoint from those
of the verifier and the reference solvers, so a defect in one grader is not inherited by the other;
none of this grading code is shipped in the public tree.

\textbf{Scalable reward design.} Because the latent world is defined before task generation, every correct intermediate state can be rendered directly and evaluated by $c_v$. This provides exact and deterministic progress feedback for every generated seed and scale without the human step labels used by process reward models~\citep{lightman2023verify}, a trained reward model, or a language-model judge.

\subsection{Progress-Potential Reward Shaping (PPRS)}
\vspace{-2mm}
\label{sec:dense:shaping}

Let $s_0,\dots,s_T$ be the workspace states along a rollout $\tau$, with $s_0$ the untouched public
tree, so $\Phi(s_0)=0$. As in potential-based reward shaping~\citep[PBRS;][]{ng1999policy}, the per-step shaping
reward is the change in potential,
$r_t=\Phi(s_t)-\Phi(s_{t-1})$, and the rollout reward is
\vspace{-2mm}
\begin{equation}
R(\tau)\;=\;\mathds{1}[\tau\ \text{passes}]\;+\;\lambda\sum_{t=1}^{T} r_t
\;=\;\mathds{1}[\tau\ \text{passes}]\;+\;\lambda\,\Phi(s_T),
\label{eq:shaped}
\end{equation}
with $\lambda=0.25$ in every run (Appendix~\ref{app:train:recipe}). Unlike PBRS, which zeroes the terminal potential so that optimal policies are unchanged, PPRS keeps $\lambda\Phi(s_T)$ and deliberately changes the objective (Appendix~\ref{app:dense}). Submission is an action, not a
state: $\Phi$ grades workspaces, and the terminal term grades the workspace the policy submits. A
rollout that exhausts its turn or token budget without submitting is not graded terminally and earns
$\lambda\Phi(s_T)$ alone; this is how the complete but unsubmitted rollout of Figure~\ref{fig:potential}
reaches $\Phi=1$ with terminal reward $0$. Because $\Phi$ is a function of the workspace alone,
$R(\tau)$ depends on the rollout only through its final state $s_T$: two rollouts that reach the same
workspace by different routes, in a different order of steps or with a different implementation,
receive the same reward.
The telescoping form also keeps the terminal reward's ranking at the top: since $\Phi\le 1$, any passing rollout earns $1+\lambda$, strictly
more than any failing one, so no single rollout is rewarded for trading completion against partial
progress; for stochastic policies, Appendix~\ref{app:dense} bounds when the preference can reverse. Appendix~\ref{app:dense} states what else is and is not preserved.

Substituting Eq.~\eqref{eq:shaped} into Eq.~\eqref{eq:grpo-adv} shows why this repairs the
starvation of GRPO. A group in which no rollout passes now has rewards
$R_i=\lambda\Phi(s_T^{(i)})$, where $s_T^{(i)}$ is the final workspace of $\tau_i$. These differ
whenever the rollouts progressed to different depths, so the group's advantages are non-zero and rank the rollouts by progress; Figure~\ref{fig:potential} shows one such
group, and Appendix~\ref{app:dense} the 24 groups we probed.

\begin{table}[t]
\vspace{-8mm}

\centering
\setlength{\tabcolsep}{1pt}
\renewcommand{\arraystretch}{1.1}
\caption{\textbf{Main results on \evalsuite, computing and math science domains of Agents' Last Exam (ALE).} Each cell reports the mean
graded score (\textbf{S@1}) and the fraction of tasks fully solved (\textbf{P@1}), averaged over
5 rollouts per task, together with their best-of-5 counterparts (\textbf{S@5}, \textbf{P@5}).
}
\vspace{-3mm}
\label{tab:main_results}
\resizebox{\textwidth}{!}{%
\begin{tabular}{ll cccc cccc cccc | cccc}
\toprule
\multirow{2}{*}{Model} & \multirow{2}{*}{Harness}
& \multicolumn{4}{c}{Math \& Science} & \multicolumn{4}{c}{ML}
& \multicolumn{4}{c}{Coding \& System} & \multicolumn{4}{c}{All (Avg)} \\
\cmidrule(lr){3-6}\cmidrule(lr){7-10}\cmidrule(lr){11-14}\cmidrule(lr){15-18}
& & S@1 & S@5 & P@1 & P@5 & S@1 & S@5 & P@1 & P@5
& S@1 & S@5 & P@1 & P@5 & S@1 & S@5 & P@1 & P@5 \\
\midrule
\multicolumn{18}{c}{\textit{Frontier API models}} \\
\midrule
GPT-6 Astra & Codex
 & 0.58 & 0.58 & 0.38 & 0.38 & 0.48 & 0.49 & 0.20 & 0.20 & 0.81 & 0.85 & 0.20 & 0.20 & 0.62 & 0.63 & 0.28 & 0.28 \\
Claude Fable 5 & Claude Code
 & 0.47 & 0.51 & 0.38 & 0.38 & 0.52 & 0.57 & 0.24 & 0.40 & 0.83 & 0.90 & 0.20 & 0.20 & 0.58 & 0.63 & 0.29 & 0.33 \\
GPT-5.6 (sol) & Codex
 & 0.50 & 0.51 & 0.38 & 0.38 & 0.46 & 0.49 & 0.20 & 0.20 & 0.82 & 0.85 & 0.20 & 0.20 & 0.58 & 0.60 & 0.28 & 0.28 \\

GPT-5.6 (terra) & Codex
 & 0.51 & 0.56 & 0.38 & 0.38 & 0.25 & 0.26 & 0.00 & 0.00 & 0.75 & 0.83 & 0.20 & 0.20 & 0.50 & 0.55 & 0.22 & 0.22 \\

Claude Opus 5 & Claude Code
 & 0.51 & 0.53 & 0.38 & 0.38 & 0.43 & 0.46 & 0.20 & 0.20 & 0.85 & 0.87 & 0.20 & 0.20 & 0.58 & 0.60 & 0.28 & 0.28 \\

Claude Sonnet 5 & Claude Code
 & 0.50 & 0.51 & 0.38 & 0.38 & 0.39 & 0.56 & 0.12 & 0.40 & 0.80 & 0.87 & 0.20 & 0.20 & 0.55 & 0.62 & 0.26 & 0.33 \\

Claude Opus 4.8 & Claude Code
 & 0.48 & 0.51 & 0.38 & 0.38 & 0.31 & 0.33 & 0.00 & 0.00 & 0.80 & 0.86 & 0.20 & 0.20 & 0.52 & 0.55 & 0.22 & 0.22 \\

Gemini 3.7 Flash & Gemini CLI
 & 0.48 & 0.50 & 0.38 & 0.38 & 0.26 & 0.29 & 0.00 & 0.00 & 0.84 & 0.86 & 0.20 & 0.20 & 0.52 & 0.54 & 0.22 & 0.22 \\

Gemini 3.6 Flash & Gemini CLI
 & 0.48 & 0.51 & 0.35 & 0.38 & 0.25 & 0.26 & 0.00 & 0.00 & 0.81 & 0.86 & 0.20 & 0.20 & 0.51 & 0.54 & 0.21 & 0.22 \\
 Gemini 3.1 Pro & Gemini CLI
 & 0.49 & 0.52 & 0.35 & 0.38 & 0.27 & 0.29 & 0.00 & 0.00 & 0.82 & 0.88 & 0.20 & 0.20 & 0.52 & 0.56 & 0.21 & 0.22 \\
\midrule
\multicolumn{18}{c}{\textit{Open-source models}} \\
\midrule
 Qwen3.8-2.4T-A95B & Vanillux2
 & 0.20 & 0.38 & 0.20 & 0.38 & 0.10 & 0.20 & 0.04 & 0.20 & 0.74 & 0.86 & 0.16 & 0.20 & 0.32 & 0.46 & 0.14 & 0.28 \\
  DeepSeek-V4-pro & Vanillux2
 & 0.12 & 0.25 & 0.12 & 0.25 & 0.10 & 0.20 & 0.04 & 0.20 & 0.73 & 0.88 & 0.16 & 0.20 & 0.29 & 0.41 & 0.11 & 0.22 \\
 \midrule
Qwen3.5-9B (base) & Vanillux2
 & 0.07 & 0.12 & 0.00 & 0.00 & 0.19 & 0.26 & 0.00 & 0.00 & \textbf{0.59} & 0.73 & 0.20 & 0.20 & 0.25 & 0.33 & 0.06 & 0.06 \\
\rowcolor{oursrow}
\model-9B (ours) & Vanillux2
 & \textbf{0.14} & \textbf{0.29} & \textbf{0.04} & \textbf{0.12} & \textbf{0.23} & 0.26 & 0.00 & 0.00 & 0.56 & 0.73 & 0.20 & 0.20 & \textbf{0.29} & \textbf{0.40} & \textbf{0.07} & \textbf{0.11} \\
 \bf \textcolor{deltaBlue}{$\Delta_{\text{\model-9B - Qwen3.5-9B }}$} & \multicolumn{1}{c}{$-$} & \textcolor{deltaBlue}{0.07$\uparrow$} & \textcolor{deltaBlue}{0.17$\uparrow$} & \textcolor{deltaBlue}{0.04$\uparrow$} & \textcolor{deltaBlue}{0.12$\uparrow$} & \textcolor{deltaBlue}{0.04$\uparrow$} & $-$ & $-$ & $-$ & \textcolor{black}{0.03$\downarrow$} & $-$ & $-$ & $-$ & \textcolor{deltaBlue}{0.04$\uparrow$} & \textcolor{deltaBlue}{0.07$\uparrow$} & \textcolor{deltaBlue}{0.01$\uparrow$} & \textcolor{deltaBlue}{0.05$\uparrow$}  \\

\midrule

Qwen3.8-27B (base) & Vanillux2
 & 0.15 & 0.25 & 0.15 & 0.25 & 0.21 & \textbf{0.32} & 0.00 & 0.00 & 0.74 & 0.82 & 0.20 & 0.20 & 0.33 & 0.43 & 0.12 & 0.17 \\
\rowcolor{oursrow}
\model-27B (ours) & Vanillux2
 & \textbf{0.23} & \textbf{0.50} & \textbf{0.23} & \textbf{0.50} & \textbf{0.26} & 0.29 & 0.00 & 0.00 & \textbf{0.82} & \textbf{0.86} & 0.20 & 0.20 & \textbf{0.40} & \textbf{0.54} & \textbf{0.16} & \textbf{0.28} \\
\bf \textcolor{deltaBlue}{$\Delta_{\text{\model-27B - Qwen3.8-27B }}$} & \multicolumn{1}{c}{$-$} & \textcolor{deltaBlue}{0.08$\uparrow$} & \textcolor{deltaBlue}{0.25$\uparrow$} & \textcolor{deltaBlue}{0.08$\uparrow$} & \textcolor{deltaBlue}{0.25$\uparrow$} & \textcolor{deltaBlue}{0.05$\uparrow$} & \textcolor{black}{0.03$\downarrow$} & $-$ & $-$ & \textcolor{deltaBlue}{0.08$\uparrow$} & \textcolor{deltaBlue}{0.04$\uparrow$} & $-$ & $-$ & \textcolor{deltaBlue}{0.07$\uparrow$} & \textcolor{deltaBlue}{0.11$\uparrow$} & \textcolor{deltaBlue}{0.04$\uparrow$} & \textcolor{deltaBlue}{0.11$\uparrow$} \\

 \bottomrule
\end{tabular}%
}
\vspace{-3mm}
\end{table}

\section{Experiments}
\label{sec:experiments}
\vspace{-2mm}

Our main experiments ask three questions. Does RL on
\dataset improve open models on the long-horizon tasks it was built for, do the gains generalize to
tasks of even longer horizon, and does the model keep its general coding competence on a standard
mid-horizon benchmark (\S\ref{sec:exp:main})? Finally, \S\ref{sec:exp:ablation} ablates our latent-world-first data and the dense reward approaches, separately.

\subsection{Setup}
\vspace{-2mm}
\label{sec:exp:setup}

\textbf{Training.} We train Qwen3.5-9B and Qwen3.8-27B using DPPO~\citep{qi2026dppo} with the reward in our Eq.~\eqref{eq:shaped}, scaling the training data together with the model: the 9B model trains on a 264-task subset of \dataset from the \emph{unit} difficulty tier, which is better suited for learning at this model scale, while the 27B model trains on the full corpus. We choose $G{=}24$ rollouts for each of 8 tasks per step, a response budget of
65{,}536 tokens per rollout (16{,}384 per turn), and at most 64 tool calls, in a sandbox that exposes
a single \texttt{bash} tool.  Our 27B model is trained on 4 nodes of B200 GPUs or 8 nodes of A100 GPUs for about 4 days. For more details, refer to Appendix~\ref{app:train}. To test whether the recipe transfers to a larger base model, we also train GLM-5.3-Flash with the same recipe, obtaining \model-320B, which we evaluate on the TB~3.0 and 4.0 subsets.

\textbf{Evaluation.} Long-horizon agent evaluation tasks are scarce and costly, with each requiring up to eight hours to evaluate. Our in-domain suite, \evalsuite, is the computing-and-math split of Agents'
Last Exam~\citep{sun2026agentslastexam}, listed in Appendix~\ref{app:eval:tasks}
(Table~\ref{tab:ale-cm-tasks}). On
TB~3.0 and~4.0~\citep{merrill2026terminalbench,marten2026tb3,marten2026tb4} we evaluate the ``CM'' subsets (22 and 18
tasks). To form them we asked Claude to judge, for every TB~3.0 task, whether it belongs to the
computing-and-math domain, even if only slightly; the
selection and the full task lists are given in Appendix~\ref{app:eval:tasks} (Table~\ref{tab:tb-cm-tasks}). We compare leaderboard model performance on the full ALE, TB~3.0, and TB~4.0 benchmarks with performance on our corresponding subsets. The results show that our splits are representative of full-set performance, with Pearson correlations of $r=0.96$ for TB~3.0 P@1 and $r=0.97$ for TB~4.0 P@1 (Appendix~\ref{app:eval:subset}; Tables~\ref{tab:tb3-subset-vs-full}, \ref{tab:tb4-subset-vs-full}, and~\ref{tab:ale-subset-vs-full}).

\textbf{Train/eval decontamination.} We measure lexical leakage per evaluation set: no environment of \dataset shares a 13-gram or an 8-gram with any task of \evalsuite, TB~3.0,
TB~4.0 or TB~2.1, and the median nearest-neighbor prompt Jaccard of the evaluation tasks to the dataset is
$0.13$ on TB~3.0/4.0 and $0.22$ on \evalsuite,
against $0.24$--$0.25$ and $0.28$ for tasks of the same benchmark compared with one another. For more details, refer to Appendix~\ref{app:data:decontamination},
Table~\ref{tab:decontam}.

\subsection{Main Results}
\vspace{-2mm}
\label{sec:exp:main}

\textbf{Results on long-horizon tasks.} As shown in Table~\ref{tab:main_results}, \model-9B improves mean S@1 by 16\% and nearly doubles its best-of-5 solve rate over Qwen3.5-9B, while \model-27B produces $33\%$ more full solutions than Qwen3.8-27B. Trained on the full corpus rather than the unit tier, \model-27B shows larger absolute gains than \model-9B on every metric and surpasses Qwen3.8-2.4T-A95B under the same harness on P@1. Although Claude Fable 5 reaches only $0.33$ P@5, training raises the 27B model from $0.17$ to $0.28$, reaching $85\%$ of the frontier score and closing more than two-thirds of the original gap.

\begin{table}[t]
\vspace{-8mm}

\centering
\setlength{\tabcolsep}{1pt}
\renewcommand{\arraystretch}{1.1}
\caption{\textbf{Terminal-Bench (TB) 3.0 and 4.0 ``CM'' subsets}.
Leaderboard rows are from the official per-trial results (5 trials per task);
$^{*}$results evaluated by us, not on the official leaderboard.
}

\vspace{-3mm}
\label{tab:tbench}
\resizebox{\textwidth}{!}{%
\begin{tabular}{l cc cc cc cc | cc !{\vrule width 0.4pt} cc cc cc cc | cc}
\toprule
\multirow{3}{*}{Model}
& \multicolumn{10}{c!{\vrule width 0.4pt}}{\textbf{TB 3.0} (22 tasks)}
& \multicolumn{10}{c}{\textbf{TB 4.0} (18 tasks)} \\
\cmidrule(lr){2-11}\cmidrule(lr){12-21}
& \multicolumn{2}{c}{Software} & \multicolumn{2}{c}{ML} & \multicolumn{2}{c}{Operations}
& \multicolumn{2}{c}{Security} & \multicolumn{2}{c!{\vrule width 0.4pt}}{All}
& \multicolumn{2}{c}{Software} & \multicolumn{2}{c}{ML} & \multicolumn{2}{c}{Operations}
& \multicolumn{2}{c}{Security} & \multicolumn{2}{c}{All} \\
\cmidrule(lr){2-3}\cmidrule(lr){4-5}\cmidrule(lr){6-7}\cmidrule(lr){8-9}\cmidrule(lr){10-11}
\cmidrule(lr){12-13}\cmidrule(lr){14-15}\cmidrule(lr){16-17}\cmidrule(lr){18-19}\cmidrule(lr){20-21}
& P@1 & P@5 & P@1 & P@5 & P@1 & P@5 & P@1 & P@5 & P@1 & P@5
& P@1 & P@5 & P@1 & P@5 & P@1 & P@5 & P@1 & P@5 & P@1 & P@5 \\
\midrule
Claude Fable 5 & 0.50 & 0.80 & 0.45 & 0.50 & 0.15 & 0.25 & 0.40 & 0.75 & 0.41 & 0.64 & 0.53 & 0.62 & 0.60 & 0.75 & 0.05 & 0.25 & 0.70 & 1.00 & 0.46 & 0.61 \\
Claude Opus 5 & 0.54 & 0.60 & 0.60 & 0.75 & 0.40 & 0.50 & 0.55 & 0.75 & 0.53 & 0.64 & 0.47 & 0.62 & 0.75 & 1.00 & 0.20 & 0.50 & 0.90 & 1.00 & 0.52 & 0.72 \\
GPT-5.6 (sol) & 0.32 & 0.50 & 0.65 & 0.75 & 0.10 & 0.25 & 0.70 & 0.75 & 0.41 & 0.55 & 0.33 & 0.62 & 0.70 & 1.00 & 0.00 & 0.00 & 0.80 & 1.00 & 0.39 & 0.61 \\
GPT-5.6 (terra) & 0.22 & 0.50 & 0.55 & 0.75 & 0.05 & 0.25 & 0.50 & 0.75 & 0.30 & 0.55 & 0.10 & 0.38 & 0.50 & 0.75 & 0.00 & 0.00 & 0.70 & 1.00 & 0.23 & 0.44 \\
GPT-5.6 (luna) & 0.18 & 0.30 & 0.25 & 0.50 & 0.00 & 0.00 & 0.30 & 0.50 & 0.18 & 0.32 & 0.12 & 0.25 & 0.25 & 0.50 & 0.05 & 0.25 & 0.50 & 0.50 & 0.18 & 0.33 \\
Grok 4.6 & 0.24 & 0.50 & 0.50 & 0.75 & 0.00 & 0.00 & 0.55 & 0.75 & 0.30 & 0.50 & 0.30 & 0.50 & 0.55 & 0.75 & 0.00 & 0.00 & 0.50 & 1.00 & 0.31 & 0.50 \\
Claude Sonnet 5 & 0.18 & 0.40 & 0.25 & 0.50 & 0.00 & 0.00 & 0.20 & 0.25 & 0.16 & 0.32 & 0.28 & 0.62 & 0.20 & 0.75 & 0.00 & 0.00 & 0.20 & 0.50 & 0.19 & 0.50 \\
Claude Opus 4.8 & 0.20 & 0.70 & 0.10 & 0.25 & 0.05 & 0.25 & 0.40 & 0.50 & 0.19 & 0.50 & 0.20 & 0.50 & 0.35 & 0.50 & 0.05 & 0.25 & 0.80 & 1.00 & 0.27 & 0.50 \\
Grok 4.5 & 0.14 & 0.30 & 0.45 & 1.00 & 0.15 & 0.50 & 0.50 & 0.75 & 0.26 & 0.55 & 0.12 & 0.12 & 0.35 & 0.75 & 0.00 & 0.00 & 0.60 & 1.00 & 0.20 & 0.33 \\
GLM 5.2 & 0.04 & 0.20 & 0.05 & 0.25 & 0.00 & 0.00 & 0.05 & 0.25 & 0.04 & 0.18 & 0.06$^{*}$ & 0.24$^{*}$ & 0.02$^{*}$ & 0.10$^{*}$ & 0.00$^{*}$ & 0.00$^{*}$ & 0.00$^{*}$ & 0.00$^{*}$ & 0.03$^{*}$ & 0.13$^{*}$ \\
Gemini 3.7 Flash$^{*}$ & 0.08 & 0.20 & 0.00 & 0.00 & 0.00 & 0.00 & 0.20 & 0.50 & 0.07 & 0.18 & 0.12 & 0.25 & 0.00 & 0.00 & 0.00 & 0.00 & 0.30 & 0.50 & 0.09 & 0.17 \\
Gemini 3.6 Flash$^{*}$ & 0.04 & 0.20 & 0.00 & 0.00 & 0.00 & 0.00 & 0.00 & 0.00 & 0.02 & 0.09 & 0.20 & 0.25 & 0.00 & 0.00 & 0.00 & 0.00 & 0.00 & 0.00 & 0.09 & 0.11 \\
\midrule
Qwen3.8-27B (base) & \textbf{0.17} & \textbf{0.43} & 0.00 & 0.00 & 0.00 & 0.00 & 0.05 & 0.25 & 0.09 & 0.24 & 0.07 & 0.31 & 0.05 & 0.25 & 0.00 & 0.00 & 0.00 & 0.00 & 0.04 & 0.19 \\
\rowcolor{oursrow}
\model-27B  & 0.15 & 0.39 & \textbf{0.19} & \textbf{0.25} & 0.00 & 0.00 & \textbf{0.20} & 0.25 & \textbf{0.14} & \textbf{0.27} & \textbf{0.17} & \textbf{0.33} & \textbf{0.12} & \textbf{0.46} & 0.00 & 0.00 & 0.00 & 0.00 & \textbf{0.10} & \textbf{0.25} \\

\bf \textcolor{deltaBlue}{$\Delta_{\text{\model-27B - Qwen3.8-27B }}$} & \textcolor{black}{0.02$\downarrow$} & \textcolor{black}{0.04$\downarrow$} & \textcolor{deltaBlue}{0.19$\uparrow$} & \textcolor{deltaBlue}{0.25$\uparrow$} & $-$ & $-$ & \textcolor{deltaBlue}{0.15$\uparrow$} & $-$ & \textcolor{deltaBlue}{0.05$\uparrow$} & \textcolor{deltaBlue}{0.03$\uparrow$} & \textcolor{deltaBlue}{0.10$\uparrow$} & \textcolor{deltaBlue}{0.02$\uparrow$} & \textcolor{deltaBlue}{0.07$\uparrow$} & \textcolor{deltaBlue}{0.21$\uparrow$} & $-$ & $-$ & $-$ & $-$ & \textcolor{deltaBlue}{0.06$\uparrow$} & \textcolor{deltaBlue}{0.06$\uparrow$} \\

\midrule

GLM-5.3-Flash (base) & 0.18 & 0.60 & 0.35 & 0.50 & 0.05 & 0.25 & 0.22 & 0.50 & 0.20 & 0.50 & 0.16 & 0.50 & 0.60 & 0.75 & 0.05 & 0.25 & 0.68 & 1.00 & 0.29 & 0.56 \\
\rowcolor{oursrow}
\model-320B & \textbf{0.21} & 0.70 & \textbf{0.40} & 0.50 & \textbf{0.10} & 0.25 & \textbf{0.45} & 0.50 & \textbf{0.27} & 0.55 & \textbf{0.25} & 0.50 & \textbf{0.65} & 0.75 & \textbf{0.15} & 0.25 & \textbf{0.70} & 1.00 & \textbf{0.37} & 0.56 \\

\bf \textcolor{deltaBlue}{$\Delta_{\text{\model-320B - GLM-5.3-Flash }}$} & \textcolor{deltaBlue}{0.03$\uparrow$} & \textcolor{deltaBlue}{0.10$\uparrow$} & \textcolor{deltaBlue}{0.05$\uparrow$} & $-$ & \textcolor{deltaBlue}{0.05$\uparrow$} & $-$ & \textcolor{deltaBlue}{0.23$\uparrow$} & $-$ & \textcolor{deltaBlue}{0.07$\uparrow$} & \textcolor{deltaBlue}{0.05$\uparrow$} & \textcolor{deltaBlue}{0.09$\uparrow$} & $-$ & \textcolor{deltaBlue}{0.05$\uparrow$} & $-$ & \textcolor{deltaBlue}{0.10$\uparrow$} & $-$ & \textcolor{deltaBlue}{0.02$\uparrow$} & $-$ & \textcolor{deltaBlue}{0.08$\uparrow$} & $-$ \\

\bottomrule
\end{tabular}%
}
\vspace{-3mm}
\end{table}

\begin{wraptable}[15]{r}{0.5\textwidth}
\vspace{-\intextsep}
\centering
\scriptsize
\setlength{\tabcolsep}{9pt}
\renewcommand{\arraystretch}{1.1}
\caption{\textbf{Terminal-Bench (TB) 2.1 results.} Frontier-reference rows are taken from the
official TB 2.1 leaderboard (5 trials, 89 tasks).
}
\vspace{-3mm}
\label{tab:tb21}
\begin{tabular}{lcc}
\toprule
Model & Pass@1 & Pass@5 \\
\midrule
GPT-6 Astra (Codex, high)    & 0.874 & 0.944 \\
Fable 5 (Claude Code, xhigh) & 0.838 & 0.933 \\
Opus 4.8 (Claude Code, high) & 0.789 & 0.944 \\
Sonnet 5 (Claude Code, high) & 0.746 & 0.921 \\
\midrule
Qwen3.5-9B (base)  & 0.178 & 0.292 \\
\rowcolor{oursrow}
\model-9B (ours)  & \textbf{0.227} & \textbf{0.348} \\
\bf \textcolor{deltaBlue}{$\Delta_{\text{\model-9B - Qwen3.5-9B }}$} & \textcolor{deltaBlue}{0.049$\uparrow$} & \textcolor{deltaBlue}{0.056$\uparrow$} \\
\midrule
Qwen3.8-27B (base) & 0.764 & 0.864 \\
\rowcolor{oursrow}
\model-27B (ours) & \textbf{0.794} & \textbf{0.884} \\
\bf \textcolor{deltaBlue}{$\Delta_{\text{\model-27B - Qwen3.8-27B }}$} & \textcolor{deltaBlue}{0.030$\uparrow$} & \textcolor{deltaBlue}{0.020$\uparrow$} \\
\bottomrule
\end{tabular}
\end{wraptable}
\textbf{Results on longer-horizon tasks.}  The TB 3.0 and 4.0 ``CM'' subsets are even more challenging, with each task requiring eight hours of expert work. Our 27B base model fully solves only $9\%$ and $4\%$ of attempted tasks, respectively. As shown in Table~\ref{tab:tbench}, \model-27B raises
P@1 from $0.04$ to $0.10$ ($2.5\times$) on TB 4.0 and from $0.09$ to $0.14$ ($+56\%$) on TB 3.0.
Notably, \model-27B surpasses GLM~5.2, the strongest open-source model on the leaderboard.
Compared with frontier models, however, a substantial gap remains. Claude Opus 5, the strongest agent on both suites, achieves P@1 scores of $0.53$ on TB~3.0 and $0.52$ on TB~4.0, suggesting that stronger open-source base models are needed to fully benefit from our high-quality but challenging \dataset.

\textbf{Scaling the base model.} Consistent with this, the recipe carries over to a stronger base. Applying it to GLM-5.3-Flash yields \model-320B, whose P@1 rises from $0.20$ to $0.27$ on TB~3.0 and from $0.29$ to $0.37$ on TB~4.0, with gains in every category on both suites and the largest on TB~3.0 security ($0.22$ to $0.45$). The improvement is concentrated in P@1: P@5 gains $0.05$ on TB~3.0 and is unchanged on TB~4.0, so training mainly makes the model solve the tasks it can already reach more reliably rather than unlocking new ones. \model-320B is the strongest open-source model on both subsets; on TB~3.0 it edges past Grok~4.5 and Claude Opus~4.8, and on TB~4.0 it trails only Claude Opus~5, Claude Fable~5, and GPT-5.6 (sol) in P@1 (Figure~\ref{fig:tb4-ranking}).

\textbf{Results on middle-horizon tasks.} We next evaluate our model on TB 2.1 (Table~\ref{tab:tb21}), a standard mid-horizon terminal benchmark. Despite zero data targeting TB 2.1, our model shows strong \textit{zero-shot} performance. \model-9B significantly outperforms its base after training, raising P@1 by $28\%$ and P@5 by $19\%$, showing that training on long-horizon tasks also improves performance on standard mid-horizon work. Our 27B model achieves P@1 $0.794$ and P@5 $0.884$, a smaller but consistent gain over its base and on par with Claude Opus 4.8, though still behind Claude Fable 5 and GPT-6 Astra.

\textbf{Qualitative results.} Figure~\ref{fig:qual-example} traces the key decision points of four rollouts on the \texttt{k8s\_payment\_api\_root\_cause\_analysis} task: every agent finds the same root causes, but \model-27B and Claude Fable~5 ground the liveness-probe cause in the measurement that explains it and judge rollback safe, whereas the base model and GPT-6 Astra cite only the symptoms and decline rollback, forfeiting the decisive grader component. For more qualitative examples, see Appendix~\ref{app:eval:qualitative}.

\subsection{Ablation Study}
\vspace{-2mm}
\label{sec:exp:ablation}

Due to resource constraints, we conduct the ablation study in the 9B setting of \S\ref{sec:exp:setup}.

\textbf{Ablation on data curation methods.} Our pipeline differs from standard LLM-synthesized corpora in two key ways: tasks are derived from a preexisting latent world and its code-compiled workflow DAG, and verifiers evaluate against this hidden ground truth rather than only agent-visible files. Table~\ref{tab:ablation} shows which components each ablation retains, with GPT-5.6 (sol, xhigh) replacing the others. \emph{Public-view verifier} ablates only access to hidden ground truth by asking the model to write a verifier from the agent's view. \emph{Prompt-first}, a standard self-instruct baseline, instead replaces the entire construction pipeline by generating each prompt before its inputs, verifier, and solution, leaving no latent world or DAG. The public-view variant causes 88\% of sampled groups to be discarded during training. Because both variants use terminal rewards, we compare them with the \emph{Sparse} row: neither improves over the base model in S@1 ($0.24$ and $0.23$ vs.\ $0.25$), whereas our full method leads them by $0.05$--$0.06$ in S@1 and $0.05$ in P@5.

\textbf{Ablation on dense reward.} Holding \dataset fixed, we compare our PPRS, which uses return-to-go $G_t$ with $\gamma=1$ and normalizes advantages by turn index, against terminal-only sparse reward, PPRS with per-rollout normalization (\emph{pooled norm.}), and direct additive reward using $\Phi(s_t)$ with $\gamma=0$~\citep{yuan2026vpr}. Our method provides the best balance of performance and training efficiency, improving S@1 by $0.04$ over sparse reward. Turn-index normalization preserves credit for later turns by avoiding comparison of their smaller returns against the whole-rollout mean, while $\gamma=1$ propagates progress back to earlier actions. In contrast, direct additive reward repeatedly rewards high-potential states without propagating credit, leading to more rollouts exhausting the context window before submission and producing longer trajectories. The advantage over sparse reward widens further at 27B, where shaped reward trains to completion but terminal-only reward produces almost entirely zero-advantage groups and fails to complete a single optimizer step.

\begin{table}[t]
\vspace{-8mm}

\centering
\setlength{\tabcolsep}{3pt}
\renewcommand{\arraystretch}{1.1}
\caption{\textbf{Ablations on \evalsuite (Qwen3.5-9B).}
Here \emph{GPU-h/step} is the mean time of one training step times the 32 GPUs of the recipe in H100 GPU-hours.
}
\vspace{-3mm}
\label{tab:ablation}
\resizebox{\textwidth}{!}{%
\begin{tabular}{l ccc lll c cccc}
\toprule
\multirow{2}{*}{Method} & \multicolumn{3}{c}{Data} & \multicolumn{3}{c}{Reward}
& \multirow{2}{*}{\shortstack{GPU-h\\/ step}} & \multicolumn{4}{c}{\evalsuite} \\
\cmidrule(lr){2-4}\cmidrule(lr){5-7}\cmidrule(lr){9-12}
& World & DAG & Hidden truth & Signal & Credit & Adv.\ norm.
& & S@1 & S@5 & P@1 & P@5 \\
\midrule
Qwen3.5-9B (base) & -- & -- & -- & no RL & -- & --
 & -- & 0.25 & 0.33 & 0.06 & 0.06 \\
\rowcolor{oursrow}
\model-9B (ours) & \ding{51} & \ding{51} & \ding{51} & $\Delta\Phi_t$ + terminal & $G_t$, $\gamma{=}1$ & per turn
 & \textbf{9.0} & \textbf{0.29} & \textbf{0.40} & \textbf{0.07} & \textbf{0.11} \\
\midrule
\multicolumn{12}{l}{\textit{Data ablation: replace one construction step with an LLM; terminal reward, compare with Sparse}} \\
\midrule
Public-view verifier & \ding{51} & \ding{51} & LLM & terminal & whole rollout & per rollout
 & 22.3 & 0.24 & 0.35 & 0.04 & 0.06 \\
Prompt-first & LLM & LLM & LLM & terminal & whole rollout & per rollout
 & 11.8 & 0.23 & 0.32 & 0.06 & 0.06 \\
\midrule
\multicolumn{12}{l}{\textit{Reward ablation: keep \dataset, change how $\Phi$ enters training}} \\
\midrule
Sparse (terminal only) & \ding{51} & \ding{51} & \ding{51} & terminal & whole rollout & per rollout
 & 16.7 & 0.25 & 0.32 & 0.06 & 0.06 \\
PPRS, pooled norm. & \ding{51} & \ding{51} & \ding{51} & $\Delta\Phi_t$ + terminal & $G_t$, $\gamma{=}1$ & per rollout
 & 10.9 & 0.28 & 0.35 & 0.06 & 0.06 \\
Direct reward (additive) & \ding{51} & \ding{51} & \ding{51} & $\Phi(s_t)$ + terminal & $G_t$, $\gamma{=}0$ & per turn
 & 9.7 & \textbf{0.29} & 0.35 & 0.06 & 0.06 \\
\bottomrule
\end{tabular}%
}
\vspace{-3mm}
\end{table}
\suppressfloats[t]

\newcommand{\qstep}[2]{{\raggedright{\color{light-gray}\tiny t#1}\;#2\par}\vspace{1.4pt}}
\newcommand{\qkey}[3]{\colorbox{#1}{\parbox{\dimexpr\linewidth-2\fboxsep\relax}{\raggedright{\color{light-gray}\tiny t#2}\;#3}}\par\vspace{1.6pt}}
\newcommand{\qdots}{{\centering\color{light-gray}$\cdots$\par}\vspace{0.6pt}}
\newcommand{\qhead}[2]{\textbf{#1}\hfill score: \textbf{#2}\par\vspace{1.5pt}\hrule\vspace{3pt}}

\begin{figure}[t]
\vspace{-8mm}
\centering
\scriptsize
\setlength{\fboxsep}{2pt}
\begin{minipage}[t]{0.238\textwidth}
\qhead{Claude Fable 5}{0.97}
\qstep{3}{Plans: read, write, validate \dots}
\qstep{4--10}{Reads prompt, cluster state, YAML, pod log \dots}
\qkey{columngreen}{14}{Writes the report \dots\ probe cause quotes \texttt{initialDelaySeconds: 5} \dots\ \texttt{cache warm-up complete (duration=6.67s)} \dots\ ``takes roughly 8\,s to start \dots\ probe fires before the server is listening'' \dots\ \texttt{safe\_to\_rollback: true} \dots\ ``pod-template settings of revision 17, not \dots\ schema changes.''}
\qstep{18--21}{Verifies every quote is verbatim \dots\ summarizes the incident \dots\ submits.}
\end{minipage}\hfill
\begin{minipage}[t]{0.238\textwidth}
\qhead{GPT-6 Astra}{0.75}
\qstep{3}{Plans: ``inspect the evidence, write the RCA, and validate its quotes'' \dots}
\qstep{6}{Dumps cluster state and YAML in one shell call \dots\ checks for \texttt{apply\_patch} \dots}
\qkey{columnred}{8}{``OOM kills are confirmed. HPA values conflict; rollback safety is unverified'' $\to$ rollback verdict: \emph{not safe} \dots}
\qkey{columnred}{9}{Writes the report \dots\ probe cause cites the settings and failures but not the warm-up measurement \dots\ not credited.}
\qstep{11--13}{Validates schema, 41 quotes \dots\ submits: ``\dots\ why rollback safety remains unverified.''}
\end{minipage}\hfill
\begin{minipage}[t]{0.238\textwidth}
\qhead{Qwen3.8-27B base}{0.70}
\qstep{2}{Lists the task directory \dots}
\qstep{4--10}{Reads the four inputs \dots}
\qstep{12--16}{Checks the em-dash and column spacing byte by byte \dots\ writes \texttt{build\_rca.py} to extract exact lines \dots}
\qkey{columnred}{18}{Builds the report \dots\ probe cause: ``the app needs ${\sim}$6.6\,s \dots\ for cache warm-up \dots'' but evidence = probe-failure lines $+$ YAML block only \dots\ ``Do NOT roll back \dots\ it would OOM/crash-loop in the same way'' $\to$ \texttt{safe\_to\_rollback: false}.}
\qstep{20--24}{Independent re-verification: validity, schema, 48 quotes verbatim \dots\ submits.}
\end{minipage}\hfill
\begin{minipage}[t]{0.238\textwidth}
\qhead{\model-27B (ours)}{0.97}
\qstep{2--18}{Reads the four inputs \dots\ confirms planned quotes byte for byte \dots}
\qkey{columngreen}{20}{Probe cause: \texttt{initialDelaySeconds: 5} \dots\ \texttt{cache warm-up complete (duration=6.67s \dots)} \dots\ ``warm-up \dots\ longer than the 5\,s initial delay'' \dots}
\qstep{28}{Verifies 29 quotes \dots}
\qkey{columngreen}{30}{Reconsiders rollback: ``\dots\ `started flapping after a routine chart-value change' \dots\ roll back first'' $\to$ \texttt{safe\_to\_rollback: true}.}
\qstep{32--40}{Re-validates \dots\ submits.}
\end{minipage}
\vspace{-1mm}
\caption{\textbf{Qualitative example on the \evalsuite task: k8s payment api root cause analysis} (Coding \& System). From a
\texttt{kubectl} capture, the Deployment/HPA YAML and a crashing pod's log, the agent writes a JSON root-cause report with \emph{verbatim} evidence strings and a rollback verdict.
\colorbox{columngreen}{Green} marks a decision that earns a component,
\colorbox{columnred}{red} one that forfeits it.
}
\label{fig:qual-example}
\vspace{-3mm}
\end{figure}

\section{Related Work}
\label{sec:related}
\vspace{-2mm}

\textbf{Synthetic corpora for coding and terminal agents.} Recent corpora typically mine repositories for issues and tests or use language models to generate tasks such as bug-injected repositories and reconstructed terminal sessions~\citep{jimenez2024swebench,yang2025swesmith,ivison2026tmax,lin2026cligym}. Other work scales executable tool-use environments and their success conditions~\citep{cai2025autoforge,fang2025agentscaler}, although scaling helps only when difficulty is controlled~\citep{zhu2026beyondscaling}. In contrast, our environments contain long, multi-stage workflows across mathematics and science, machine learning, coding, and systems; are instantiated by seeded generator code rather than directly authored by a language model; and provide a dense progress signal for every task.

\textbf{Backward construction.} Deriving a task from a known solution is an established family:
denoising, controlled error injection, code mutation, environment rewinding, instruction
back-translation, and procedural puzzle generation all corrupt or invert a single artifact whose
ground truth is already
known~\citep{lewis2019bart,arocena2015bart,tran2013geco,yang2025swesmith,hariharan2025breakpoint,lin2026cligym,li2023backtranslation,wu2024inversecoder,stojanovski2025reasoninggym,chen2025enigmata,liu2025synlogic,li2025internbootcamp}.
We instead corrupt a rendering of an entire latent world, so the task, verifier, and progress
signal are all functions of that one world rather than separately generated factors that can drift
apart~\citep{zeng2026ace,ivanov2026anchor}.

\textbf{Sparse reward and dense signal in agentic RL.} Existing methods avoid uninformative rollout groups by filtering identical rewards, selecting tasks near the policy's current ability, or adding expert prefixes~\citep{yu2025dapo,florensa2018goalgan,zhang2025bread}.
Other approaches provide denser feedback through learned judges~\citep{cui2025prime,choudhury2025agentprm} or proxy signals such as test-pass rates~\citep{du2025swedev,shen2026seta,yang2026t1}, but these require additional models or capture only terminal outcomes.
Our potential $\Phi$ instead measures exact progress and regression against the latent world at every workflow stage, without requiring a critic or language-model judge.
For full related work, please refer to Appendix~\ref{app:related}.

\section{Conclusion}
\label{sec:conclusion}
\vspace{-2mm}

We introduce a latent-world-first approach for scaling verifiable long-horizon agent training. Each task is built backward from a known-correct latent world, which is rendered into interrelated files, modified through controlled corruptions, and reused for exact verification and dense progress rewards. We instantiate this approach in \dataset, a collection of 1{,}170 verified environments across seven domains, and show that training improves both 9B and 27B open models on ALE-CM, with gains transferring to Terminal-Bench tasks of different horizons and to a 320B base model.
One limitation of our current study is its focus on synthetic terminal environments, and the hardest benchmarks still show a gap between our models and leading frontier agents. Supporting new workflows also requires authoring and validating additional environment components. Future work could simplify this process, expand domain coverage, and combine the resulting data with stronger open models, making more professional workflows available for reproducible RL without human step labels or language-model judges.

\subsection*{Acknowledgments}
\vspace{-2mm}

We thank Scott Cohen at Adobe Research for mentorship, guidance, and support throughout this
project, and Adobe Research for supporting this work. We are grateful to our colleagues at Adobe
Research and UCLA for helpful discussions and feedback. 

\subsection*{AI use statement}
\vspace{-2mm}

We used generative AI tools for the following tasks that require disclosure. \emph{Generating
synthetic data and implementing methods:} coding agents helped author the generators of \dataset,
that is, the environment specifications and their reference solvers, strict verifiers, and progress
scorers (\S\ref{sec:data}; Appendix~\ref{app:data:determinism}). Once authored, a generator is a
deterministic seeded program, so no language model participates when an environment is generated,
verified, or rewarded; every generator passed the qualification gates of
Appendix~\ref{app:data:gates}, and the authors reviewed each specification. \emph{Qualitative data
analysis:} Claude helped judge which TB~3.0 and~4.0 tasks relate to the computing-and-math domain of
our training data, which defines our ``CM'' evaluation subsets; Appendix~\ref{app:eval:tasks} gives the procedure and the
resulting task lists, and Appendix~\ref{app:eval:subset} checks that the subsets track
full-benchmark performance. Coding assistants also helped write evaluation and analysis scripts.
Separately, language models appear in our experiments as objects of study rather than as aids to
the authors: GPT-5.6 (sol) builds the two data-ablation baselines (\S\ref{sec:exp:ablation}),
Gemini~3.6~Flash is the fixed evaluator of the corpus difficulty comparison
(Appendix~\ref{app:data:comparison}), and frontier models are evaluated as baselines. No generative
model grades any task or produces any training reward. Among tasks with recommended disclosure, we
used generative AI tools to search and summarize related literature and to draft and edit parts of
the text. Translation, surveys, interviews, and transcription are not applicable to this work. We
have reviewed all AI-assisted work, and all claims, numbers, and citations were checked by the
authors. We take responsibility for the final content of this work, including text, claims, and
artifacts produced with the aid of generative AI.

\subsection*{Ethics statement}
\vspace{-2mm}

\dataset is fully synthetic: every environment is generated procedurally from a seed, contains no
data collected from people, and involves no human subjects, so the corpus encodes no private records
or proprietary code. Task specifications are drawn from public benchmarks and reference material
(Appendix~\ref{app:data:sources}), but every artifact an agent sees is generated by the environment
itself, and third-party benchmarks and corpora used for evaluation and comparison are used under
their published licenses (Appendix~\ref{app:data:comparison}). The environments simulate
professional workflows such as data reconciliation, checkpoint consolidation, systems
administration, and the analysis of synthetic binaries and packet captures in the style of
capture-the-flag exercises. Each declares its toolchain in its public contract, the agent's
workspace is seeded only with its public tree, and no task targets real software, systems, or people. Improving open-weight agents on
long-horizon terminal tasks carries the general dual-use considerations of more capable coding
agents; our security-flavored tasks are analysis and forensics exercises over synthetic artifacts,
the same class of task already present in public terminal-agent benchmarks. Because part of the
training mix is built against the interfaces of \evalsuite, we measure train/evaluation overlap for
every evaluation set and report it with its calibration (\S\ref{sec:exp:setup};
Appendix~\ref{app:data:decontamination}). Frontier baselines run in their native agent harnesses,
the setting most favorable to them (Appendix~\ref{app:eval}), and we discuss the limitations of
our results in \S\ref{sec:conclusion}.

\subsection*{Reproducibility statement}
\vspace{-2mm}

We have written the paper and appendix so that every result can be reproduced from them.
\emph{Data.} Every environment of \dataset is a deterministic, model-free function of its
generator, seed, and scale tier (\S\ref{sec:data:pipeline}; Appendix~\ref{app:data:determinism}),
so the corpus can be regenerated exactly, at a median of 30\,s of CPU time per environment and with
no model calls. The generation stages are specified in \S\ref{sec:data:pipeline}; task sources and
the domain taxonomy in Appendix~\ref{app:data:sources} (Table~\ref{tab:domain-sources}); the
coverage policy in Appendix~\ref{app:data:coverage}; the scale ladder and the measurement of
horizon and difficulty in Appendices~\ref{app:data:scale} and~\ref{app:data:difficulty}; the
qualification gates in Appendix~\ref{app:data:gates}
(Table~\ref{tab:data-gates}); and per-domain statistics in Appendix~\ref{app:data:stats}
(Table~\ref{tab:data-composition}). Appendix~\ref{app:data:examples} walks through three complete
training environments, each with its public view and the ground truth sealed behind it, so the
construction can be followed end to end (the listings are excerpts of the package files). \emph{Reward and training.} The progress potential and the
shaped reward are fully specified by Eqs.~\eqref{eq:potential} and~\eqref{eq:shaped}. All three base
models are publicly available and DPPO is a published objective; \S\ref{sec:exp:setup} states the
corpus each model is trained on, and Table~\ref{tab:hparams} lists the hyperparameters shared by
every run. Each ablation changes a single component of this recipe. \emph{Evaluation.} All evaluation benchmarks are
public. Appendix~\ref{app:eval} gives the harnesses, turn and token budgets, number of rollouts, and
metric definitions; Appendix~\ref{app:eval:tasks} lists every task in \evalsuite and in the
TB~3.0/4.0 ``CM'' subsets (Tables~\ref{tab:ale-cm-tasks} and~\ref{tab:tb-cm-tasks}); and
Appendix~\ref{app:eval:subset} shows that the subsets track full-benchmark performance. The
difficulty comparison of Table~\ref{tab:data-comparison} follows Appendix~\ref{app:data:comparison},
which fixes the evaluator, harness, sampling budget, and subsample seed, and train/evaluation
overlap is measured with the protocol of Appendix~\ref{app:data:decontamination}.

\bibliography{custom}
\bibliographystyle{iclr2027_conference}

\appendix
\clearpage
\addtocontents{toc}{\protect\setcounter{tocdepth}{2}}
\renewcommand{\contentsname}{Appendix Contents}
\tableofcontents
\clearpage

\section{Data Construction Details}
\label{app:data}

This appendix expands the condensed description in \S\ref{sec:data}.

\subsection{Task Sources and Domain Taxonomy}
\vspace{-2mm}
\label{app:data:sources}

\dataset's generators target the seven computing-and-math subdomains of Agents' Last
Exam~\citep{sun2026agentslastexam}, which constitute \evalsuite and which we refer to throughout as
\dataset's seven domains: software engineering (9.1), data and analytics engineering (9.2), AI
engineering, safety and CS research (9.3), mathematical and operations research (9.4),
cybersecurity and digital forensics (9.5), infrastructure engineering and cloud operations (9.6),
and quantum computing (9.7). Table~\ref{tab:domain-sources} gives the scope of each and the
deterministic signals it is verified by; subdomain ids and names are taken from the ALE taxonomy
and the scope summaries follow it, reconstructed from O\textsuperscript{*}NET/SOC~2018 and the
released ALE task specifications. The coverage policy of Appendix~\ref{app:data:coverage} requires
every domain to be populated by at least two generators.

\textbf{Sourcing.} For every domain we assemble a pool of real task specifications: specifications mined from the internet, MLE-bench (82
Kaggle competitions), or, for domain~9.7 (quantum computing), which no benchmark suite covers,
public reference material (MQT Bench's circuit suite, Qiskit Experiments manuals, TeNPy's DMRG
examples). From this pool we randomly sample specifications and author our own generator around
each selection: every generator is a fresh procedural design of our own, drawing on the sampled
specification for structure and difficulty only, and never reproducing its prompt, input, answer,
or grader text. A feasibility filter first restricts candidates to specifications whose
ground truth is knowable by construction and executable batch/CPU-only, excluding live-service,
vision/audio, heavy-toolchain, GPU-kernel, and open-ended ``improve this'' tasks; a similarity gate
then compares every new generator's code, specification, and generated data with all
already-registered generators and rejects near-duplicates; and an adversarial review pass attacks
each accepted generator's specification for hidden
shortcuts, too short a horizon, or a disguised source before implementation.

\begin{table}[t]
\centering
\small
\caption{\textbf{The seven computing-and-math subdomains of \evalsuite that \dataset spans.}
Subdomains (9.1--9.7) of the Computing \& Math domain of Agents' Last
Exam~\citep{sun2026agentslastexam}, which we refer to as \dataset's seven domains. Ids and names
are taken from the ALE taxonomy; scope summaries follow it, reconstructed from
O\textsuperscript{*}NET/SOC~2018 and the released ALE task specifications. ``Verified by'' names
the deterministic signals we build the strict verifier and progress potential around.}
\label{tab:domain-sources}
\begin{tabular}{p{2.1cm} p{10.4cm}}
\toprule
Subdomain & Scope \\
\midrule
9.1 Software Engineering &
Feature implementation, debugging, refactoring, and test authoring across real codebases, turning issues and specs into working, tested code changes. \emph{Verified by:} test-suite pass (gate then score), behavioral correctness under fixed inputs, and no regression in existing tests. \\
9.2 Data \& Analytics Engineering &
ETL/ELT pipelines, recommender and ranking systems, feature engineering, and analytics workflows, turning raw data into validated datasets, features, and model outputs. \emph{Verified by:} output values exact or within tolerance, schema validity (gate), metric reproduction, and feature-parity against a reference. \\
9.3 AI Engineering, Safety \& CS Research &
Model training and evaluation pipelines, research-paper reproduction, checkpoint and config management, and alignment/safety experiments, turning specs and code into reproduced results. \emph{Verified by:} reproduced metrics within tolerance (FID, accuracy, etc.), config and alignment exactness, artifact validity, and deterministic-under-seed behavior. \\
9.4 Mathematical \& Operations Research &
Combinatorial optimization, game-theoretic equilibrium computation, applied mathematics and number theory, and probabilistic estimation and filtering, implemented from specifications. \emph{Verified by:} objective or solution exact or within tolerance, an optimality/feasibility certificate (gate), and reproducibility under a fixed seed. \\
9.5 Cybersecurity \& Digital Forensics &
Reverse engineering and malware-config extraction, network and packet triage, crackme and key-generation analysis, and incident forensics, turning binaries and captures into extracted artifacts and findings. \emph{Verified by:} extracted-value exactness against a reference (hash/string), finding-set precision and recall, and decoded-artifact correctness (gate). \\
9.6 Infrastructure Engineering \& Cloud Operations &
Kubernetes migration and configuration, root-cause analysis of production incidents, OS and security hardening, and infrastructure-as-code, turning infrastructure specs and incidents into working, secured systems. \emph{Verified by:} the system's world-state (resources/health) under fixed inputs, config correctness against a reference, incident-cause exactness, and policy-check pass (gate). \\
9.7 Quantum Computing &
Quantum-circuit construction and simulation, quantum-error-correction code analysis, and circuit-level noise simulation, turning circuit and code specs into measured outcomes and logical-error estimates. \emph{Verified by:} statistics and error-rates within tolerance against a reference, reproducibility under a fixed seed, and circuit/structure validity. \\
\bottomrule
\end{tabular}
\end{table}

\subsection{Determinism, Reproducibility, and Generation Cost}
\vspace{-2mm}
\label{app:data:determinism}

Every stage of \S\ref{sec:data:pipeline} is a pure, seeded function, and no language model
participates in generation. Each generator's code is authored once by coding agents and reviewed by
the authors (see AI use statement); running it is model-free. An environment is therefore fully
reproducible from its generator, seed, and scale tier, and two seeds of the same generator yield
provably distinct worlds (the six-seed diversity gate of Appendix~\ref{app:data:gates}). The cost
of an environment therefore splits in two. A one-time authoring cost per generator is amortized
over every seed and scale tier at which it is instantiated. The marginal cost is seconds of
compute: producing one environment, including running both reference solvers, sealing the hidden
suite, and previewing the progress potential $\Phi_W$, takes a median of 30\,s of CPU time (90th
percentile 82\,s) and no model calls, against hours of expert time per hand-authored task. Each
environment declares its toolchain (e.g., \texttt{python3}, \texttt{sqlite3}) in its public
contract, and the agent's workspace is seeded only with its public tree.

\subsection{Coverage Axes and Covering Array}
\vspace{-2mm}
\label{app:data:coverage}

Every environment is labeled on 19 axes, the required axes of the corpus coverage policy:
curriculum layer, subdomain, generator revision, scale tier, the four horizon axes (semantic, action,
operational, verification), workflow stage, tool capability, artifact role, relation type,
corruption class, source kind, license class, reward component, observed reward band, outcome
category, and failure category. The corpus must satisfy a strength-2 covering array over these
axes: for any two axes, every valid pair of their values co-occurs in at least one admitted
environment. Coverage is
extended greedily by choosing the generator, seed, and scale tier that fill the largest number of
currently uncovered pairs, so ``diverse'' is a checkable property of the corpus rather than a
claim. Pairwise coverage is what separates enforced from nominal diversity and rules out
\emph{template locking}~\citep{zeng2026ace}: a corpus can be balanced on every axis taken alone
while its axis values only ever appear in a few fixed combinations; a strength-2 covering array
rules this out for every pair of axes. Coverage is checked on admitted environments only, after the
gates of Appendix~\ref{app:data:gates}. Appendix~\ref{app:data:stats} reports the resulting balance
across domains and capability categories.

\textbf{Multi-format artifacts.} Text-only tasks under-represent real terminal work, so each
environment ships concrete, heterogeneous artifacts inspected through standard tooling: relational
databases, columnar and delimited text, serialized model shards, compiled binaries, and service
manifests. Some environments are also multimodal: part of the latent world is rendered as an image,
such as the billing-dashboard screenshot of Example~1 (Appendix~\ref{app:data:examples}).

\textbf{Per-seed world variation.} Re-seeding a generator varies entity identities, values,
layouts, tensor names, and the placement of every corruption, so no two environments share a public
tree (Appendix~\ref{app:data:gates}). This is the mechanism behind both scale and decontamination.

\subsection{Scale Ladder}
\vspace{-2mm}
\label{app:data:scale}

Each generator exposes a scale axis that tunes the size of the world. For the warehouse
environment the tiers run from 30 customers and 48 transactions per month (small) to 480 customers
and 960 transactions per month (large), with proportional growth in duplicates and missing values;
for the checkpoint-consolidation environment the ladder is the shard count crossed with the
parallelism topology (pure tensor parallelism up to combined tensor-, pipeline-, and
expert-parallel layouts). Because tasks are defined over the workflow DAG, typically five to six
critical stages with cross-stage dependencies, the action, operational, and verification horizons
grow with the tier as first-class, controllable quantities. We sample across the ladder so that the
corpus is neither uniformly trivial nor uniformly unsolvable. For the generators instantiated at all
three scale tiers, the median input tree grows from 17 files and 51\,KB (unit) to 20 files and
168\,KB (small) and 28 files and 586\,KB (full) while the workflow DAG is unchanged.

\subsection{Measuring Horizon and Difficulty}
\vspace{-2mm}
\label{app:data:difficulty}

We quote the critical path of the workflow DAG (Stage~2 of \S\ref{sec:data:pipeline}), not the
observed rollout length $T$, as the structural measure of horizon, because it does not inflate with
an agent's detours. The scale tier and the dependency depth of the DAG are generation controls, not
difficulty scores; difficulty is relative to a learner $\pi$, so we report it as verified success
under a declared harness, $p_\pi(E)=\Pr_{\tau\sim\pi(\cdot\mid E)}\bigl[\operatorname{ver}_W(s_T)=1\bigr]$.
Appendix~\ref{app:data:stats} reports this success rate for an untrained Qwen3.8-27B under the
training harness and for a fixed frontier evaluator (Table~\ref{tab:data-composition}). Where $p_\pi(E)$ saturates at zero, the progress potential
$\Phi_W(s_T)$ of \S\ref{sec:dense} still resolves difficulty; this is the regime the dense reward
is built for. Table~\ref{tab:data-comparison} positions \dataset against recent agentic corpora.

\subsection{Qualification Gates}
\vspace{-2mm}
\label{app:data:gates}

A synthesized environment is admitted only if it passes every gate below. All gates are
benchmark-independent; they are enforced while each generator is authored and checked automatically
when each environment is generated.
\begin{itemize}[leftmargin=*, itemsep=1pt, topsep=2pt]
\item \textbf{Reward causality.} Each reward component $c$ must lose value when its cause is
mutated in the workspace: $c(\mu_c(s^\star))<c(s^\star)$ for a passing workspace $s^\star$ and its
cause-mutation $\mu_c$.
\item \textbf{Exploit resistance.} A probe $s^{(k)}$ for every one of 14 gaming strategies (empty
output, minimal schema-valid output, copied input, single-seed hard-coding, path traversal, private
lookup, and others) must score $0$ under both the strict verifier and the progress scorer,
$\operatorname{ver}_W(s^{(k)})=\Phi_W(s^{(k)})=0$.
\item \textbf{Role independence.} Reference solver, alternative solver, strict verifier, and
progress scorer have disjoint, audited code closures; none imports another.
\item \textbf{Solver agreement.} The reference solver $\pi^{\mathrm{ref}}$, whose output scores
exactly $1.0$, $\operatorname{ver}_W(\pi^{\mathrm{ref}}(s_0))=\Phi_W(\pi^{\mathrm{ref}}(s_0))=1$,
and the independently authored alternative solver $\pi^{\mathrm{alt}}$ must produce identical output
trees on every seed, $\pi^{\mathrm{ref}}(s_0)=\pi^{\mathrm{alt}}(s_0)$; a disagreement is treated
as an unresolved specification and the environment is rejected.
\item \textbf{Progress validity.} The potential $\Phi_W$ satisfies $0<\Phi_W(s)<1$ on partially
correct states, is monotone to $1.0$ along the reference trajectory, and produces negative
increments on regression.
\item \textbf{Metamorphic invariance.} Semantics-preserving transforms of a correct output do not
change the grade; stale-output counterfactuals fail.
\item \textbf{Six-seed diversity.} At least six seeds yield distinct world identities, public-tree
hashes, and task ids.
\item \textbf{Public-contract hygiene.} No private authority, reference output, or verifier code
appears in the prompt $x$ or the public tree $s_0$.
\end{itemize}

\textbf{Why we gate at generation time.} The solver audit is independent of the verifier: neither
$\pi^{\mathrm{ref}}$ nor $\pi^{\mathrm{alt}}$ imports verifier code, so their agreement is evidence
about the specification rather than about the grader. Unlike pipelines that skip validation and
rely on downstream RL to soft-filter unsolvable tasks~\citep{ivison2026tmax}, we gate at generation
time, because a dense progress signal is only useful if it is causal and resists gaming.
Procedural seeding also makes decontamination structural: no two environments share a public tree,
$s_0(E)\neq s_0(E')$ for $E\neq E'$, and none shares one with an evaluation instance
(Appendix~\ref{app:data:decontamination}).

\begin{table}[t]
\centering
\small
\caption{\textbf{Intrinsic qualification gates.} Every task environment must pass all gates; each
gate is benchmark-independent, enforced while each generator is authored, and checked automatically
when each environment is generated. The third column names the failure class each gate is matched
to; passing one gate is not taken as evidence for a property it does not observe.}
\vspace{-3mm}
\label{tab:data-gates}
\resizebox{\textwidth}{!}{%
\begin{tabular}{lll}
\toprule
Gate & What it enforces & Failure class it targets \\
\midrule
Reward causality        & each reward component drops when its cause is mutated & reward credits the wrong outcome \\
Exploit resistance      & all 14 gaming probes score $0$                        & verifier loophole, shortcut acceptance \\
Role independence       & disjoint reference/alt/verifier/progress closures     & shared bias between solver and grader \\
Solver agreement        & reference and alternative solvers yield identical trees & verifier-only acceptance, unresolved spec. \\
Progress validity       & $0<\Phi<1$; monotone; decreases on regression         & invalid process signal \\
Metamorphic invariance  & meaning-preserving rewrites keep the grade; stale outputs fail & over-narrow verifier; stale-state acceptance \\
Six-seed diversity      & distinct world / public-tree / task id ($\geq 6$)     & degenerate environment, template repetition \\
Public-contract hygiene & no private authority in the prompt or public tree    & answer leakage \\
\bottomrule
\end{tabular}%
}
\vspace{-2mm}
\end{table}

\subsection{Dataset Statistics}
\vspace{-2mm}
\label{app:data:stats}

\begin{table}[t]
\centering
\small
\caption{\textbf{Composition of \dataset by capability category.} Counts are task environments;
horizon columns are per-environment medians over the
category (workflow stages, input files the agent sees, injected corruptions); ``formats'' counts
distinct wire formats among the inputs. Base P@1 is the single-rollout pass rate of untrained
Qwen3.8-27B under the training harness, from one probe on our environments; frontier P@1
is the fixed evaluator of Appendix~\ref{app:data:comparison} on its 92-environment subsample.}
\label{tab:data-composition}
\resizebox{\textwidth}{!}{%
\begin{tabular}{l c cccc cc}
\toprule
& & \multicolumn{4}{c}{Per-environment horizon (median)} & \multicolumn{2}{c}{Difficulty (P@1)} \\
\cmidrule(lr){3-6}\cmidrule(lr){7-8}
Category & Task envs. & Stages & Input files & Corruptions & Formats & Base 27B & Frontier \\
\midrule
Math \& Science      & 264 & 6 & 8 & 1 & 7 & 0.04 & 0.31 \\
ML                   & 242 & 6 & 12 & 1 & 11 & 0.03 & 0.24 \\
Coding \& System     & 664 & 6 & 14 & 3 & 19 & 0.03 & 0.30 \\
\midrule
\rowcolor{oursrow}
\textbf{\dataset (total)} & 1{,}170 & 6 & 12 & 2 & 15 & 0.03 & 0.29\\
\bottomrule
\end{tabular}%
}
\vspace{-5mm}
\end{table}

\textbf{Balance.} We quantify balance with the normalized-entropy score
$\mathrm{Balance}=\exp(H)/K$, where $H=-\sum_{i=1}^{K} p_i \log p_i$ and $p_i$ is the fraction of
environments in bucket $i$ of $K$; the score lies in $[1/K,1]$ with $1$ perfectly uniform. \dataset
attains 0.89 across the three capability categories and 0.91 across the seven domains.
\textbf{Horizon.} An
environment has a median of 6 workflow stages (range 3--12), 8 progress nodes,
6 reward components, and 2 injected corruptions (1--7), and exposes to the agent a median of
12 input files (up to 501) totaling 51\,KB (up to 6.1\,MB) in
23 wire formats (32 file extensions), described by a task prompt of 583 tokens at the median
(Qwen tokenizer; maximum 944). Under the training harness of \S\ref{sec:exp:setup}, an untrained
Qwen3.8-27B uses a median of 19 tool calls and generates 30k tokens per rollout, and 52\% of its
rollouts exhaust the 65{,}536-token context before submitting.

\textbf{Difficulty.} The same rollouts give the base-model difficulty: across our single-rollout
probes (one per environment), 3.2\% of environments are fully solved (mean graded score 0.06).
Terminal reward alone is therefore near zero on most of the corpus, which is the regime
\S\ref{sec:dense} addresses. Under the fixed evaluator of Appendix~\ref{app:data:comparison},
Gemini~3.6~Flash reaches pass@1 29\% and pass@8 40\% on a fixed 92-environment subsample
(Table~\ref{tab:data-comparison}), so the open--frontier gap is roughly $3\%$ against $29\%$; on 26
of those 92 environments, and on 52 further environments probed at the small and full scale tiers,
it solves nothing and submits an answer in at most 29\% of its rollouts, so a substantial part of
\dataset is beyond current frontier reach as authored.

\subsection{Comparison Protocol}
\vspace{-2mm}
\label{app:data:comparison}

\textbf{Agent time.} ``Agent time'' in Table~\ref{tab:data-comparison} is measured rather than
estimated. The fixed evaluator of the difficulty protocol below, Gemini~3.6~Flash, is run on a fixed
100-task subsample of each corpus, and for every rollout we record the wall-clock time from the
agent's first action to its last step. The interval includes every model call and retry and every
command execution in between; it excludes only sandbox setup before the first action and the
separate test run after the last step (\dataset's host-side grade at submission, 0.3\,s at the
median, falls inside the last step). The table reports the median and mean over rollouts, in
minutes. For the prior corpora these are the rollouts of the difficulty measurement below (Harbor
\texttt{terminus-2}, 60 turns, 8 rollouts per task, timed by Harbor's \texttt{agent\_execution}
interval). That budget ends only 1--13\% of their rollouts, and every rollout it ends has already
run longer than its corpus's median, so the reported prior-corpus medians equal those of an unlimited
budget (only the means could rise). \dataset's rollouts run longer (28\% use more than 60 tool
calls), so for \dataset we lift the budget until it no longer binds: we draw a fixed
100-environment subsample from all 1{,}170 environments (seed \texttt{20260915}, stratified by
domain 9.1--9.7) and run the evaluator in the single-\texttt{bash} harness of \S\ref{sec:exp:setup}
with a budget of 300 tool calls and 2\,h of agent wall-clock per rollout, 2 rollouts per
environment. No rollout reaches either limit (median 41 tool calls, maximum 145), so the values are
not truncated: median 12.4\,min and mean 14.8\,min, against a median of 0.5--2.0\,min for the prior
corpora. The measure compares corpora under one evaluator and is not an estimate of
human effort. Terminal-Universe releases no environments and cannot be timed.
``Progressive reward'' marks a training signal defined on
intermediate workspace states during an episode (\S\ref{sec:dense}); SETA's test fraction is computed
once at the end of the episode and therefore does not qualify.

\textbf{Difficulty.} We measure difficulty uniformly: a single evaluator, \textbf{Gemini~3.6~Flash},
is run on a fixed subsample of every corpus at an identical budget (60 tool calls, 8 rollouts per
task, temperature~1.0), and we report pass@1 (mean over rollouts) and the unbiased pass@8 (macro over
tasks). For \dataset the evaluator runs in the single-\texttt{bash} harness of \S\ref{sec:exp:setup}
on a fixed 92-environment subsample, reaching pass@1 29\% and
pass@8 40\%; it solves nothing on 26 of the 92, nor on the 52 further environments probed at the
small and full tiers (Appendix~\ref{app:data:stats}). For the prior corpora we run the same evaluator
under the official Harbor \texttt{terminus-2} agent on a fixed 100-task subsample of each (seed
\texttt{20260915}, stratified by the corpus's own domain or repository field where present):
SETA 80/88 (\texttt{camel-ai/seta-env-release}, CC~BY-SA~4.0; 100/100 tasks), TMAX-15K 74/80
(pre-converted Harbor release of \texttt{allenai/TMax-15K}, ODC-BY; 100/100), Endless Terminals 83/89
(\texttt{obiwan96/endless-terminals}, MIT; 99/100), SWE-Smith 47/52 (\texttt{SWE-bench/SWE-smith},
MIT; 100/100), and CLI-Gym 69/74 (\texttt{LiberCoders/CLI-Gym}, MIT; 89/100). Excluded are tasks the
Docker sandbox cannot faithfully build: 11 CLI-Gym and 1 Endless tasks whose released Dockerfiles
reference build-context files absent from the release or perform build-time \texttt{/etc/hosts} or
\texttt{/proc} writes (these are CLI-Gym's hardest systems-level corruptions, so its number is a mild
upper bound); one TMAX task is under-sampled. Every exclusion is recorded. Terminal-Universe is a
task-generation method whose 32k environments are not publicly released, so its difficulty could not
be measured and its cells are marked NA. The evaluator and budget are shared across all rows; only
the scaffold differs (prior corpora \texttt{terminus-2}, \dataset single-\texttt{bash}). Sizes are as
reported by each corpus; ``oracle'' names what grades a task and ``grading metric'' what it emits to a
learner, and a corpus is credited with a progressive reward only if it emits more than a terminal scalar.

Table~\ref{tab:data-axes} complements the size and difficulty comparison with the axes on which
generation pipelines differ in design: what anchors generation, whether a model is in the loop,
what admits a task and when, whether difficulty and coverage are controlled or only reported, and
what a learner receives short of a terminal pass.

\begin{table}[t]
\centering
\small
\caption{\textbf{How recent agentic corpora are generated and gated.} Axes follow the ACE
lens~\citep{zeng2026ace}. \emph{Anchor} is the factor from which the others are derived;
\emph{model in generation} lists what a language model writes; \emph{validation} is what admits a
task and when; \emph{difficulty} and \emph{coverage} record whether the corpus controls them (``--''
marks no explicit control reported); \emph{progress signal} is what a learner receives short of a
terminal pass. Prior-corpus cells are transcribed from the cited papers.}
\vspace{-3mm}
\label{tab:data-axes}
\resizebox{\textwidth}{!}{%
\begin{tabular}{l l l l l l l}
\toprule
Corpus & Anchor & Model in generation & Validation & Difficulty control & Coverage control & Progress signal \\
\midrule
SWE-Smith~\citep{yang2025swesmith}               & repository        & bug injection, task text  & tests run post hoc      & -- & -- & none \\
Terminal-Universe~\citep{wu2026terminaluniverse} & agent trajectory  & trajectory, task text     & pytest post hoc         & -- & -- & none \\
TMAX-15K~\citep{ivison2026tmax}                  & task text         & task, container, tests    & none; RL soft-filters   & -- & topic balancing & none \\
SETA~\citep{shen2026seta}                        & task text         & task, tests               & solvability post hoc    & -- & -- & test fraction \\
Endless Terminals~\citep{gandhi2026endless}      & task text         & task, container, tests    & solvability post hoc    & -- & -- & none \\
CLI-Gym~\citep{lin2026cligym}                    & agent trajectory  & trajectory, task text     & tests post hoc          & -- & -- & none \\
\midrule
\rowcolor{oursrow}
\textbf{\dataset (ours)} & latent world & generator code (once, reviewed) & 8 gates at generation & scale $\times$ DAG depth & strength-2 covering array & staged $\Phi$ \\
\bottomrule
\end{tabular}%
}
\vspace{-5mm}
\end{table}

\subsection{Decontamination}
\vspace{-2mm}
\label{app:data:decontamination}

\textbf{Structural cleanliness.} Because each task is projected from a freshly seeded procedural
world, no two environments, and in particular no environment and no evaluation instance, share a public
tree; the six-seed diversity gate enforces this by hash. \dataset therefore contains no verbatim
copy of any evaluation task.

\textbf{Measured overlap.} We additionally measure lexical overlap between the task prompt of every
one of the 1{,}170 environments and our evaluation sets with a 13-gram sliding window over lower-cased
alphanumeric tokens, following standard contamination protocol. The evaluation texts are the task
cards of the 18 computing-and-math tasks (and, as a superset, all 23 tasks of that split; 9{,}010
distinct 13-grams), the instructions of the 74 TB 3.0/4.0 tasks (21{,}138) and of the 89
TB 2.1 tasks (13{,}292). No environment shares a single 13-gram with any evaluation task (0 of
1{,}170 against each set; maximum shared fraction 0.0\%), and none shares an 8-gram; overlap appears
only at 5-grams, in generic phrasing such as ``not modify files under input''. A control confirms
that the comparator is sensitive: the two TB sets share 156 13-grams with each other.

\textbf{Per-set breakdown.} Table~\ref{tab:decontam} gives the per-set view behind these statements,
with TB~3.0 and TB~4.0 separated (full sets and the CM subsets of Table~\ref{tab:tbench}) and TB~2.1
included, and adds three statistics to the 13-gram count. \emph{8-gram hits} repeats the count at the
shorter window; \emph{max token coverage} is, for the worst evaluation task, the fraction of its tokens
that fall inside an 8-gram also present in some environment prompt; \emph{NN Jaccard} is each evaluation
task's largest token-set Jaccard similarity to any of the 1{,}170 prompts, summarized by its median and
maximum over tasks. The calibration block applies the same code to a task scored against the other tasks of its own
benchmark, which shares a 13-gram
in every TB case (the common closing instruction) and reaches a nearest-neighbor Jaccard of
$0.37$. Against that scale, no evaluation task shares an 8-gram with any environment, and
nearest-neighbor similarity to the whole dataset stays at $0.13$ (median) and $0.18$ (maximum) on
TB~3.0 and 4.0 and $0.10$ and $0.16$ on TB~2.1; on \evalsuite it is $0.22$ and $0.38$.

\begin{table}[t]
\centering
\small
\setlength{\tabcolsep}{5pt}
\renewcommand{\arraystretch}{1.1}
\caption{\textbf{Lexical overlap between all 1{,}170 environments of \dataset and each evaluation set.}
\emph{Hits}: evaluation tasks sharing at least one 13-gram (resp.\ 8-gram) of lower-cased alphanumeric
tokens with any environment prompt. \emph{Max token cov.}: for the worst evaluation task, the share of
its tokens covered by 8-grams that also occur in some environment prompt. \emph{NN Jaccard}: per
evaluation task, the largest token-set Jaccard similarity to any environment prompt; median\,/\,max over
tasks. CM = the 22-\,/\,18-task computing-and-math subsets evaluated in
Table~\ref{tab:tbench}. The calibration block applies the same code to same-benchmark, different-task pairs.}
\vspace{-2mm}
\label{tab:decontam}
\resizebox{\textwidth}{!}{%
\begin{tabular}{@{}lrrrrc@{}}
\toprule
Evaluation set & Tasks & 13-gram hits & 8-gram hits & Max token cov. & NN Jaccard med.\,/\,max \\
\midrule
\evalsuite & 18 & 0 & 0 & 0.0\% & 0.22 / 0.38 \\
TB~3.0, all & 74 & 0 & 0 & 0.0\% & 0.13 / 0.18 \\
TB~3.0, CM subset & 22 & 0 & 0 & 0.0\% & 0.14 / 0.18 \\
TB~4.0, all & 66 & 0 & 0 & 0.0\% & 0.13 / 0.18 \\
TB~4.0, CM subset & 18 & 0 & 0 & 0.0\% & 0.14 / 0.18 \\
TB~2.1 & 89 & 0 & 0 & 0.0\% & 0.10 / 0.16 \\
\midrule
\multicolumn{6}{@{}l}{\textit{Calibration: same code, same-benchmark tasks}} \\
\evalsuite task vs.\ other \evalsuite tasks & 18 & 2 & 13 & 26.0\% & 0.28 / 0.34 \\
TB~3.0 CM task vs.\ other TB~3.0 tasks & 22 & 22 & 22 & 42.5\% & 0.24 / 0.37 \\
TB~4.0 CM task vs.\ other TB~4.0 tasks & 18 & 18 & 18 & 42.5\% & 0.25 / 0.37 \\
\bottomrule
\end{tabular}}
\end{table}

\subsection{Three Worked Training Environments}
\vspace{-2mm}
\label{app:data:examples}

\lstdefinestyle{alefile}{basicstyle=\ttfamily\scriptsize,breaklines=true,columns=fullflexible,keepspaces=true,frame=single,framerule=0.3pt,rulecolor=\color{light-gray},xleftmargin=3pt,xrightmargin=3pt,aboveskip=3pt,belowskip=6pt,showstringspaces=false,postbreak=\mbox{\textcolor{light-gray}{$\hookrightarrow$}\space}}
\lstdefinestyle{aletext}{style=alefile,postbreak={},breakindent=0pt}

Two generalizations of \S\ref{sec:data:pipeline} are easiest to see on concrete environments.
First, the public tree $s_0$ need not be text: Example~1, a cloud-operations task from one of
\dataset's seven domains, stages a billing-dashboard screenshot whose three findings appear in no
CSV. Second, nothing in the pipeline is specific to the seven
computing-and-math domains: Examples~2 and~3 were synthesized by the same pipeline for two other
domains of Agents' Last Exam~\citep{sun2026agentslastexam}, energy engineering (closed-loop
building control) and accounting (financial-statement reconstruction). For each we present the
training environment only, in two parts: the \emph{public view}, the prompt $x$ and the input tree
$s_0$ that are all the agent ever receives, and the \emph{ground truth} sealed behind them, the
latent world $W$, the reference deliverables derived from it, and what the grader compares. The
three also span the two ways a latent world grounds a reference: in Examples~1 and~3 the world is a
generated record set and the reference follows from it by construction; in Example~2 the world is
a pinned simulator plus a declared procedure, so the reference is a \emph{tool oracle}, produced by
running that procedure once inside the sealed image. Paths are relative to the agent's working
directory, the agent writes its deliverables to \texttt{output/}, and all three run in a pinned
container with the network disabled.

\subsubsection*{Example 1: multimodal cloud cost optimization (Cloud Operations)}

\begin{figure}[H]
\centering
\includegraphics[width=0.82\textwidth]{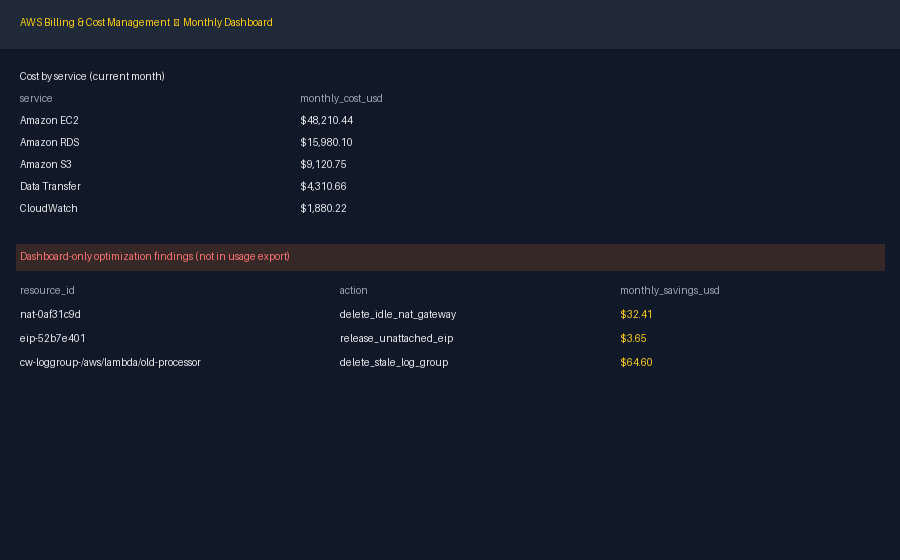}
\caption{\textbf{The dashboard screenshot staged as \texttt{input/aws\_billing\_dashboard.png} in
Example~1} ($900\times560$ px, drawn by the generator from the latent world). The upper table is
per-service context and plays no role in grading; the lower table holds the three dashboard-only
findings that appear in no CSV, so the agent must read the image to recover their ids, actions, and
savings.}
\label{fig:ex-dashboard}
\vspace{-3mm}
\end{figure}

\textbf{Public view.} The container provides \texttt{python3} and \texttt{pandas}. The agent
receives the prompt below and three inputs: an inventory CSV of 430 EC2, RDS, and S3 resources
with utilization percentiles, a pricing table, and a rendered billing-dashboard screenshot
(Figure~\ref{fig:ex-dashboard}) whose lower panel lists three findings, an idle NAT gateway, an
unattached Elastic IP, and a stale CloudWatch log group, that appear in no CSV. This is an
extra-large instance of its generator.

\begin{lstlisting}[style=aletext]
# AWS cost optimization analysis (430 inventory resources)

You are an AWS cost optimization analyst working on Linux. Three inputs are staged under `input/`:

- `input/aws_billing_dashboard.png` - a synthetic AWS billing dashboard screenshot. It shows a per-service cost breakdown AND three 'dashboard-only' findings (an idle NAT gateway, an unattached Elastic IP, and a stale CloudWatch log group) that are NOT present in the usage CSV. Inspect the image directly to recover them.
- `input/resource_usage.csv` - the EC2/RDS/S3 inventory with columns `resource_id,resource_type,name,state,instance_type,cpu_p95,mem_p95,connections_p95,size_gb,ia_access_ratio`.
- `input/aws_pricing_reference.csv` - the deterministic pricing table (`sku,dimension,unit,price_usd`).

Apply these rules and price every saving from the pricing table:
- Downsize a running EC2 instance one tier when `cpu_p95 < 20` and `mem_p95 < 50`.
- Downsize an available RDS instance one tier when `cpu_p95 < 25` and `connections_p95 < 10`.
- Move an S3 bucket to Infrequent Access when `ia_access_ratio < 0.10` and `size_gb >= 100`.
- Include the three dashboard-only findings (delete the idle NAT gateway, release the unattached Elastic IP, delete the stale CloudWatch log group).

Write `output/optimization_report.json` with an `executive_summary` object and a `recommendations` array, and `output/savings_summary.csv` with columns `resource_id,resource_type,name,action,current_cost,projected_cost,monthly_savings`.

Scoring is weighted: 30% coverage F1 over the hidden recommendation set (including the three dashboard-only resources), 30% action correctness, 25% savings accuracy (full credit within +/-20% relative error), and 15% report completeness.
\end{lstlisting}

\begin{lstlisting}[style=alefile]
input/
  aws_billing_dashboard.png  900 x 560 px dashboard render: cost by service + 3 findings
  resource_usage.csv         430 rows: 220 EC2 instances, 80 RDS instances, 130 S3 buckets
  aws_pricing_reference.csv  11 SKUs: 5 EC2 tiers, 4 RDS tiers, 2 S3 storage classes
\end{lstlisting}

\noindent The two CSVs are plain tables; the inventory leaves the columns that do not apply to a
resource type empty.

\begin{lstlisting}[style=alefile]
# input/aws_pricing_reference.csv (complete)
sku,dimension,unit,price_usd
t3.medium,ec2_compute,hour,0.041600
m5.large,ec2_compute,hour,0.096000
m5.xlarge,ec2_compute,hour,0.192000
m5.2xlarge,ec2_compute,hour,0.384000
m5.4xlarge,ec2_compute,hour,0.768000
db.t3.medium,rds_compute,hour,0.068000
db.m5.large,rds_compute,hour,0.171000
db.m5.xlarge,rds_compute,hour,0.342000
db.m5.2xlarge,rds_compute,hour,0.684000
s3-standard,s3_storage,gb_month,0.023000
s3-standard-ia,s3_storage,gb_month,0.012500

# input/resource_usage.csv (6 of 430 rows)
resource_id,resource_type,name,state,instance_type,cpu_p95,mem_p95,connections_p95,size_gb,ia_access_ratio
db-000067,rds,rds-svc-067,available,db.m5.xlarge,62.12,,346,,
db-000048,rds,rds-svc-048,available,db.m5.xlarge,18.48,,8,,
i-000161,ec2,ec2-app-161,running,m5.2xlarge,45.55,74.81,,,
i-000000,ec2,ec2-app-000,running,m5.2xlarge,11.97,41.34,,,
s3-000015,s3,data-bucket-015,,,,,,5923,0.052
s3-000118,s3,data-bucket-118,,,,,,19,0.596
\end{lstlisting}

\textbf{Ground truth.} The latent world $W$ is the generator's record of the fleet: the 430
inventory rows with their utilization percentiles and sizes, the pricing tiers in their declared
order, and three planted dashboard-only findings with their dashboard-reported costs. Rendering
splits $W$ across three formats: the inventory becomes \texttt{resource\_usage.csv}, the
tiers become \texttt{aws\_pricing\_reference.csv}, and the three findings are drawn only into the
PNG, next to a per-service cost table that is context, so that no text file contains them. The
reference follows from $W$ by applying the prompt's four rules with the pricing table: 730 hours per
month, one tier down along the declared order, S3 Standard to Infrequent Access per GB-month, and
recommendations sorted by savings descending, then by id. Each recommendation's current cost,
projected cost, and savings are computed from unrounded prices and rounded to cents separately, so a
row's savings can differ by one cent from the difference of its two rounded costs (\texttt{s3-000126},
7,734\,GB: $177.88-96.68=81.20$, savings 81.21). Of the six rows
above, three qualify: \texttt{db-000048} (CPU 18.48, 8 connections) drops to \texttt{db.m5.large}
and saves 124.83; \texttt{i-000000} (CPU 11.97, memory 41.34) drops to \texttt{m5.xlarge} and saves
140.16; \texttt{s3-000015} (5,923\,GB at 5.2\% infrequent access) moves to Infrequent Access and
saves 62.19. For this seed the rules yield 148 recommendations, 74 EC2 downsizes, 27 RDS downsizes,
44 S3 tier moves, and the three dashboard findings. The executive summary sums the rounded rows
column by column: current cost 32,629.73 USD, projected cost 16,234.79, and monthly savings
16,395.02. The savings total therefore exceeds current minus projected (16,394.94) by 0.08, the sum
of the one-cent residues on 12 S3 rows, far inside the grader's 20\% savings tolerance. The grader applies
two hard gates (both files parse and are non-empty; total savings lie between zero and total current
cost), then scores the weighted rubric the prompt announces: coverage F1 of recommended ids against
the hidden set (30\%), action correctness on matched ids (30\%), savings accuracy with full credit
within 20\% relative error and linear decay to zero at 100\% (25\%), and completeness of the summary
fields, per-recommendation columns, and CSV header (15\%); it also reports recall of the three
dashboard-only ids separately, accepting the CloudWatch id as printed on the dashboard as an alias
of the canonical id.

\begin{lstlisting}[style=alefile]
# reference/optimization_report.json (148 recommendations; excerpt, reformatted)
{"executive_summary": {"resource_count": 430, "recommendation_count": 148,
     "dashboard_only_finding_count": 3, "total_current_cost": 32629.73,
     "total_projected_cost": 16234.79, "total_monthly_savings": 16395.02},
 "recommendations": [
  {"resource_id": "i-000003", "resource_type": "ec2", "name": "ec2-app-003",
   "action": "downsize_ec2", "current_cost": 560.64, "projected_cost": 280.32,
   "monthly_savings": 280.32},
  {"resource_id": "i-000006", "resource_type": "ec2", "name": "ec2-app-006",
   "action": "downsize_ec2", "current_cost": 560.64, "projected_cost": 280.32,
   "monthly_savings": 280.32},
  ...
  {"resource_id": "cw-loggroup-old-processor", "resource_type": "cloudwatch_log_group",
   "name": "/aws/lambda/old-processor", "action": "delete_stale_log_group",
   "current_cost": 64.6, "projected_cost": 0.0, "monthly_savings": 64.6},
  ...
  {"resource_id": "nat-0af31c9d", "resource_type": "nat_gateway",
   "name": "nat-gw-prod-1c", "action": "delete_idle_nat_gateway",
   "current_cost": 32.41, "projected_cost": 0.0, "monthly_savings": 32.41},
  ...
  {"resource_id": "eip-52b7e401", "resource_type": "elastic_ip",
   "name": "eip-orphaned-legacy", "action": "release_unattached_eip",
   "current_cost": 3.65, "projected_cost": 0.0, "monthly_savings": 3.65},
  ...]}

# reference/savings_summary.csv (148 rows; first 3)
resource_id,resource_type,name,action,current_cost,projected_cost,monthly_savings
i-000003,ec2,ec2-app-003,downsize_ec2,560.64,280.32,280.32
i-000006,ec2,ec2-app-006,downsize_ec2,560.64,280.32,280.32
i-000024,ec2,ec2-app-024,downsize_ec2,560.64,280.32,280.32
\end{lstlisting}

\subsubsection*{Example 2: closed-loop building MPC (Energy, Power \& Nuclear Engineering)}

\textbf{Public view.} The container is the \texttt{engineering-energyplus-mpc} profile
(EnergyPlus~22.1.0, cvxpy~1.6.5, numpy~2.1.3; 4~cores, 8\,GB, 5\,h wall clock). The agent receives
the prompt below and the input tree that follows it, and nothing else. This is a medium instance of
its generator.

\begin{lstlisting}[style=aletext]
# EnergyPlus closed-loop building MPC (7 days at 15 minutes)

Use the exact toolchain declared in `input/instance.json`. Inspect `input/model.idf`, `input/weather.epw`, `input/experiment_plan.json`, and `input/output_contract.json`. The model is the pinned EnergyPlus 22.1 one-zone example with a declared internal-mass transformation and a schedule actuator driving the `MPC Cooling Sink`.

Run the deterministic thermostat baseline and supplied excitation at 15-minute resolution. Fit the declared second-order input-output thermal model over both EnergyPlus traces and preserve its training and held-out diagnostics. This is a 3R2C-comparable dynamic-order model, not a claim that physical resistance and capacitance parameters were identified.

Implement two distinct cvxpy/OSQP receding-horizon policies using the declared price, peak-load, comfort, and switching weights. Deploy each through the EnergyPlus Python API: at every zone timestep read the simulated zone temperature, solve the policy, and actuate the first control. A precomputed open-loop schedule does not satisfy this task.

Write the nine artifacts named in `input/output_contract.json`. The three data traces cover the complete simulation window. The two action files and comparison metrics cover the final 24 hours. RC logs must expose model coefficients, training and held-out calibration, policy-specific closed-loop error, objective weights, and solver status counts.

The grader densely compares all nine artifacts to a reference generated and rerun with EnergyPlus 22.1 and cvxpy in the pinned profile.
\end{lstlisting}

\begin{lstlisting}[style=alefile]
input/
  question.md           the task prompt above
  instance.json         horizon, schedules, tariff, controller weights (excerpt below)
  experiment_plan.json  four runs, identification recipe and split, two policies
  output_contract.json  the nine deliverables, their columns, and their rounding (below)
  model.idf             one-zone EnergyPlus model, materialized from the pinned example
  weather.epw           Golden, CO (NREL) TMY3 weather file, from the same pinned install
\end{lstlisting}

\noindent\texttt{instance.json} declares every quantity the procedure depends on; its seven 672-entry
schedules are elided here.

\begin{lstlisting}[style=alefile]
{
  "kind": "energyplus_second_order_closed_loop_mpc_cosimulation", "seed": 6610,
  "energyplus_version": "22.1.0", "cvxpy_version": "1.6.5", "numpy_version": "2.1.3",
  "source_focus": "energyplus_1zoneuncontrolled_22_1",
  "source_idf_sha256": "53e3fb38...", "source_epw_sha256": "65041e11...",
  "weather_window": "golden_tmy3_july",
  "run_period_start_month": 7, "run_period_start_day": 1, "run_period_days": 7,
  "timestep_minutes": 15, "horizon_steps": 672,
  "prediction_horizon_steps": 24, "reporting_window_steps": 96,
  "cooling_capacity_kw": 2.0406, "cooling_cop": 3.0, "internal_mass_area_m2": 48.0,
  "baseline_deadband_c": 0.5, "baseline_on_fraction": 0.85, "identification_ridge": 1e-08,
  "occupancy":            [0, 0, ...],           // 672 values; 1 between 08:00 and 20:00
  "is_peak_period":       [0, 0, ...],           // 672 values; 1 between 17:00 and 20:00
  "setpoint_c":           [27.5, 27.5, ...],     // 24.2 while occupied, else 27.5
  "comfort_min_c":        [17.0, 17.0, ...],     // 20.5 while occupied, else 17.0
  "comfort_max_c":        [28.0, 28.0, ...],     // 24.5 while occupied, else 28.0
  "energy_price_per_kwh": [0.061587, ...],       // 672 values in [0.058, 0.250]
  "excitation_schedule":  [0.9, 0.1, 0.9, ...],  // 672 values in {0.1, 0.9}
  "policy_objectives": {
    "energy_saving":   {"energy_weight": 0.35, "price_weight": 1.0, "peak_weight": 0.1,
      "on_peak_weight": 0.1, "comfort_weight": 2.0, "switch_weight": 0.12},
    "demand_response": {"energy_weight": 0.05, "price_weight": 0.35, "peak_weight": 5.0,
      "on_peak_weight": 2.0, "comfort_weight": 4.0, "switch_weight": 0.12}
  }
}
\end{lstlisting}

\noindent\texttt{output\_contract.json} names the nine deliverables and fixes their schemas.

\begin{lstlisting}[style=alefile]
{
  "trace_files": {"baseline": "baseline_data.csv",
                  "energy_saving": "mpc_energy_saving_data.csv",
                  "demand_response": "mpc_demand_response_data.csv"},
  "trace_columns": ["step", "hour", "cooling_w", "t_zone", "setpoint", "outdoor_temp_c",
                    "occupancy", "cooling_fraction", "electric_kw", "price_per_kwh",
                    "comfort_min_c", "comfort_max_c", "one_step_predicted_zone_temp_c"],
  "action_files": ["mpc_actions_energy_saving.csv", "mpc_actions_demand_response.csv"],
  "rc_log_files": ["rc_log_energy_saving.json", "rc_log_demand_response.json"],
  "metrics_file": "metrics_comparison.csv",
  "metrics_rows": ["baseline", "mpc_energy_saving", "mpc_demand_response"],
  "metrics_columns": ["cooling_kwh", "elec_kwh", "cost_usd", "peak_load_kw",
                      "peak_hour_avg_kw", "discomfort_dh", "calibration_rmse_c"],
  "summary_file": "results_summary.json",
  "numeric_rounding": {"csv": 8, "model_coefficients": 10, "summary": 8}
}
\end{lstlisting}

\textbf{Ground truth.} The latent world is the pinned simulator together with the declared
procedure: the EnergyPlus model, the weather file, and the schedules, weights, and identification
recipe of \texttt{instance.json} fix a unique correct value for every field of the nine
deliverables. The reference is therefore produced by executing that procedure once, inside the same
pinned image, by the sealed reference builder, which the agent never sees. The builder
(i) checks the SHA-256 of the shipped \texttt{1ZoneUncontrolled.idf} and Golden TMY3 weather file
against the values pinned in \texttt{instance.json} and applies the declared edits to the model
(run period 1--7 July, four timesteps per hour, an \texttt{InternalMass} of 48\,m$^2$, and a
\texttt{Schedule:Constant} actuator driving a $2.0406$\,kW \texttt{OtherEquipment} cooling sink);
(ii) runs EnergyPlus twice through its Python API, once under the two-position thermostat (declared
setpoint schedule, deadband 0.5\,$^\circ$C, on-fraction 0.85) and once under the prescribed 0.9/0.1
excitation schedule, reading the zone temperature and writing the actuator at every zone timestep;
(iii) fits the declared second-order input--output model
\begin{equation*}
T[t]=a_1T[t-1]+a_2T[t-2]+b_0T_{\mathrm{out}}[t]+b_1T_{\mathrm{out}}[t-1]+c_0u[t]+c_1u[t-1]+d
\end{equation*}
by ridge least squares ($\lambda=10^{-8}$) on the first 80\% of each trace with the last 20\% held
out, and asserts a negative cooling gain, $c_0+c_1<0$; (iv) runs EnergyPlus twice more in closed
loop, solving at every zone timestep a 24-step cvxpy/OSQP receding-horizon problem with the
policy's declared weights and actuating its first control; and (v) writes the nine artifacts with
the declared rounding: three 672-row traces, two 96-row action files over the final 24 hours, two
logs exposing the seven coefficients, training and held-out RMSE, closed-loop one-step RMSE,
objective weights, and solver-status counts, the $3\times7$ metrics table, and the summary. A
round-trip self-check re-fits the model and re-reads every artifact the way the grader will
(coefficients within $10^{-7}$, row counts, distinct policy weights) before the environment is
published. The grader then compares the agent's nine files with these references cell by cell (CSV)
and leaf by leaf (JSON) at a tolerance of $10^{-4}$, scoring each artifact by its fraction of
matching fields. Every artifact is a function of the closed-loop trajectories, so a precomputed
open-loop schedule, a first-order model, or a single policy cannot reproduce them.

\subsubsection*{Example 3: financial-statement reconstruction (Accounting \& Finance)}

\textbf{Public view.} The container is the \texttt{finance-foss-headless} profile (Python~3.10 only;
4~cores, 4\,GB, 30\,min wall clock). The agent receives the prompt below, a policy document, and a
filing packet for a fictitious company, Juniper Ridge Holdings plc, in the two shapes an analyst
meets in practice: SEC ``companyfacts'' JSON and inline-XBRL HTML filings. This is a medium
instance of its generator.

\begin{lstlisting}[style=aletext]
# Reconstruct and reconcile annual financial statements

Reconstruct the annual statements from the multi-document filing packet under `input/filings/` and the SEC-shaped `input/companyfacts.json`. Use `input/filing_index.json`, `input/account_mapping.csv`, and the authority and fallback rules in `input/reconstruction_policy.md`.

Write three deliverables:

- `output/financial_statements.json`, with company metadata, USD-million unit metadata, and a `periods` array containing balance sheet, income statement, and cash-flow sections.
- `output/evidence.csv`, with the exact columns `period, statement, line_item, selected_value, concept, source_file, locator, citation, authority_basis`, in period and mapping order. Every locator must resolve to the selected source value.
- `output/exceptions.json`, listing every cross-source conflict, explicit amendment restatement, accepted taxonomy alias, and missing-structured-fact fallback with its policy-based resolution.

Preserve integer USD millions. Verify the accounting, income, and cash roll-forward identities for every period before writing the files. Do not use quarterly, unaudited, pro forma, or wrong-unit evidence.
\end{lstlisting}

\begin{lstlisting}[style=alefile]
input/
  question.md                      the task prompt above
  instance.json                    2 fiscal years, 16 line items, 3 deliverables
  reconstruction_policy.md         authority and fallback rules (below)
  filing_navigation.md             amendments carry only restated rows; ids are locators
  account_mapping.csv              16 line items -> statement, label, accepted concepts
  filing_index.json                4 filings: form, accession, filing date, status (below)
  companyfacts.json                SEC-shaped facts: 16 us-gaap concepts x 3, whole USD
  filings/2025/form10-k.htm        FY2025 10-K, 16 inline-XBRL rows (fact-2025-000..015)
  filings/2025/form10-k-a.htm      FY2025 10-K/A, restates one row
  filings/2025/current-report.htm  FY2025 8-K with quarterly (Q3) figures
  filings/2024/form10-k.htm        FY2024 10-K
\end{lstlisting}

\noindent The policy document is the agent's only statement of precedence.

\begin{lstlisting}[style=aletext]
# Annual statement reconstruction policy

All output values are USD millions. Companyfacts `USD` values are whole USD and must be divided by 1,000,000. Use exact fiscal-year and period-end matches, annual `FY` facts, and Forms 10-K or 10-K/A only. Ignore quarterly, non-USD, pro forma, and unaudited supplemental facts.

Within Companyfacts, choose the latest `filed` date and then accession among all accepted concepts in `account_mapping.csv`. Within filing documents, a 10-K/A overrides only the rows it explicitly restates; otherwise retain the original 10-K row. When the latest structured fact and latest filing row disagree, the structured Companyfacts value controls. If no accepted structured fact exists, use the latest applicable filing row. Record every cross-source conflict, amendment restatement, taxonomy alias, and structured-fact fallback in `exceptions.json`. Every evidence row must cite an exact JSON pointer or HTML fragment that resolves to the selected value.
\end{lstlisting}

\noindent The excerpts below follow one line item, cash and cash equivalents for FY2025, through
the packet: under its balance-sheet concept it is reported five times with four different values.

\begin{lstlisting}[style=alefile]
# account_mapping.csv (first of 16 rows)
statement,line_item,filing_label,canonical_concept,accepted_concepts,period_type
balance_sheet,cash,Cash and cash equivalents,us-gaap:CashAndCashEquivalentsAtCarryingValue,us-gaap:CashAndCashEquivalentsAtCarryingValue|jrh:CashCashEquivalentsRestrictedCashAndRestrictedCashEquivalents,instant

# filing_index.json
{"cik": "8899426957", "company": "Juniper Ridge Holdings plc", "filings": [
  {"accession": "8899426957-25-000001", "document": "filings/2025/form10-k.htm",
   "filed": "2026-01-30", "fiscal_year": 2025, "form": "10-K", "status": "original"},
  {"accession": "8899426957-25-000002", "document": "filings/2025/form10-k-a.htm",
   "filed": "2026-02-15", "fiscal_year": 2025, "form": "10-K/A",
   "status": "amendment_explicit_rows_only"},
  {"accession": "8899426957-25-000004", "document": "filings/2025/current-report.htm",
   "filed": "2025-10-15", "fiscal_year": 2025, "form": "8-K",
   "status": "current_report_decoy"},
  {"accession": "8899426957-24-000001", "document": "filings/2024/form10-k.htm",
   "filed": "2025-01-30", "fiscal_year": 2024, "form": "10-K", "status": "original"}]}

# companyfacts.json, node /facts/us-gaap/CashAndCashEquivalentsAtCarryingValue/units/USD
[{"accn": "8899426957-25-000001", "form": "10-K", "fy": 2025, "fp": "FY",
  "end": "2025-12-31", "filed": "2026-01-30", "val": 239000000},
 {"accn": "8899426957-25-000002", "form": "10-K/A", "fy": 2025, "fp": "FY",
  "end": "2025-12-31", "filed": "2026-02-15", "val": 246000000},
 {"accn": "8899426957-24-000001", "form": "10-K", "fy": 2024, "fp": "FY",
  "end": "2024-12-31", "filed": "2025-01-30", "val": 256000000}]

# filings/2025/form10-k.htm, balance sheet ("USD in millions")
<tr><th scope="row">Cash and cash equivalents</th><td><ix:nonFraction id="fact-2025-000"
  name="us-gaap:CashAndCashEquivalentsAtCarryingValue" contextRef="FY2025" unitRef="USD"
  scale="6" format="ixt:numdotdecimal">239</ix:nonFraction></td></tr>
# filings/2025/form10-k-a.htm, the only row it restates
<tr><th scope="row">Cash and cash equivalents</th><td><ix:nonFraction
  id="amended-fact-2025-000" name="us-gaap:CashAndCashEquivalentsAtCarryingValue"
  contextRef="FY2025" unitRef="USD" scale="6" ...>243</ix:nonFraction></td></tr>
# filings/2025/current-report.htm, 8-K with a quarterly context
<tr><th scope="row">Cash and cash equivalents</th><td><ix:nonFraction
  id="current-report-fact-2025-000" name="us-gaap:CashAndCashEquivalentsAtCarryingValue"
  contextRef="Q32025" unitRef="USD" scale="6" ...>230</ix:nonFraction></td></tr>
\end{lstlisting}

\textbf{Ground truth.} The latent world $W$ is the company's true two-period statements: 16 integer
line items in USD millions for fiscal years 2025 and 2024, each ending December~31. $W$ satisfies
the three identities the prompt asks the agent to verify, assets $=$ liabilities $+$ equity, net
income $=$ revenue $-$ cost of revenue $-$ operating expenses, and ending cash $=$ beginning cash $+$
operating $+$ investing $+$ financing cash flow, as well as the ties that connect real statements:
the four asset lines sum to total assets (FY2025: $246+147+62+732=1{,}187$), balance-sheet cash
equals cash-flow ending cash in each year (246 and 256), and FY2025 beginning cash equals FY2024
ending cash (256). Rendering projects $W$ into the two public formats: each value becomes a
whole-dollar structured fact under its us-gaap concept in \texttt{companyfacts.json} and an
inline-XBRL row, in millions and with a stable fragment id, in the fiscal year's 10-K. The defect
plan then corrupts the clean packet, but only in sources that the policy rejects. The original
FY2025 10-K misstates three balance-sheet lines, in its rows and in its structured facts alike (cash
239, accounts receivable 138, other current assets 726, against 246, 147, and 732 in $W$); a 10-K/A
restates one row (cash, 239 $\to$ 243), and the restated value is still wrong; the whole FY2025 fact
set is refiled in \texttt{companyfacts.json} under the 10-K/A accession with the true values, so the
structured feed and the amendment disagree about the same restatement; an 8-K reports the same
concepts under quarterly \texttt{Q32025} contexts; the filings carry an ``unaudited supplemental
data'' section; outflows are printed in accounting parentheses, \texttt{(110)}; and each line item
admits a second, company-extension concept (\texttt{jrh:}) that this seed leaves unused. Because
the defect plan is declared, the correct resolution of each of these is known without running a
solver, and the three reference deliverables follow from $W$, the defect plan, and the policy
text. Since every defect lies in a rejected source, the values the policy selects are those of $W$,
and \texttt{financial\_statements.json} is $W$ itself. The rejected values are also the ones that
break the ties: the original 10-K's asset lines sum to 1,165 against its total assets of 1,187, and
neither 239 nor 243 matches the ending cash of 246. \texttt{exceptions.json} lists exactly the four
conflicts the defect plan created, each with the resolution the policy dictates; the amendment
entry records the filing-level choice (the 10-K/A row, 243, supersedes the original, 239), while
its selected value is the structured 246 that overrides both. \texttt{evidence.csv} holds one row
per line item and period, 32 in all, whose locator is the JSON pointer of the latest annual fact
and whose authority basis names the rule that selected it. A round-trip self-check reselects every
value from the public packet, verifies the three identities and the three ties above in every
period, and asserts that the observed exception set equals the expected one before the environment
is published. The grader compares the agent's three files with these references leaf by leaf and
cell by cell at zero tolerance.

\begin{lstlisting}[style=alefile]
# reference/financial_statements.json (USD millions; reformatted)
{"cik": "8899426957", "company": "Juniper Ridge Holdings plc", "currency": "USD",
 "unit": "millions", "periods": [
  {"period": "2025", "period_end": "2025-12-31",
   "balance_sheet": {"cash": 246, "accounts_receivable": 147, "inventory": 62,
                     "other_assets": 732, "assets": 1187, "liabilities": 598,
                     "equity": 589},
   "income_statement": {"revenue": 1852, "cost_of_revenue": 666,
                        "operating_expense": 927, "net_income": 259},
   "cash_flow": {"beginning_cash": 256, "operating_cash_flow": 365,
                 "investing_cash_flow": -110, "financing_cash_flow": -265,
                 "ending_cash": 246}},
  {"period": "2024", "period_end": "2024-12-31",
   "balance_sheet": {"cash": 256, "accounts_receivable": 73, "inventory": 30,
                     "other_assets": 546, "assets": 905, "liabilities": 508,
                     "equity": 397},
   "income_statement": {"revenue": 1583, "cost_of_revenue": 601,
                        "operating_expense": 777, "net_income": 205},
   "cash_flow": {"beginning_cash": 106, "operating_cash_flow": 333,
                 "investing_cash_flow": -146, "financing_cash_flow": -37,
                 "ending_cash": 256}}]}

# reference/exceptions.json (reformatted)
[{"kind": "cross_source_conflict", "period": "2025", "line_item": "accounts_receivable",
  "structured_value": 147, "filing_value": 138, "selected_value": 147,
  "resolution": "structured_feed_priority"},
 {"kind": "amendment_restatement", "period": "2025", "line_item": "cash",
  "original_value": 239, "amended_value": 243, "selected_value": 246,
  "resolution": "explicit_amendment_over_original"},
 {"kind": "cross_source_conflict", "period": "2025", "line_item": "cash",
  "structured_value": 246, "filing_value": 243, "selected_value": 246,
  "resolution": "structured_feed_priority"},
 {"kind": "cross_source_conflict", "period": "2025", "line_item": "other_assets",
  "structured_value": 732, "filing_value": 726, "selected_value": 732,
  "resolution": "structured_feed_priority"}]

# reference/evidence.csv (2 of 32 rows)
period,statement,line_item,selected_value,concept,source_file,locator,citation,authority_basis
2025,balance_sheet,cash,246,us-gaap:CashAndCashEquivalentsAtCarryingValue,companyfacts.json,#/facts/us-gaap/CashAndCashEquivalentsAtCarryingValue/units/USD/1,"SEC Companyfacts JSON, us-gaap:CashAndCashEquivalentsAtCarryingValue, FY 2025, companyfacts.json#/facts/us-gaap/CashAndCashEquivalentsAtCarryingValue/units/USD/1",latest_exact_period_annual_structured_fact
2024,balance_sheet,cash,256,us-gaap:CashAndCashEquivalentsAtCarryingValue,companyfacts.json,#/facts/us-gaap/CashAndCashEquivalentsAtCarryingValue/units/USD/2,"SEC Companyfacts JSON, us-gaap:CashAndCashEquivalentsAtCarryingValue, FY 2024, companyfacts.json#/facts/us-gaap/CashAndCashEquivalentsAtCarryingValue/units/USD/2",latest_exact_period_annual_structured_fact
\end{lstlisting}

\section{Dense Reward Details}
\label{app:dense}

This appendix expands \S\ref{sec:dense}.

\textbf{Wasted groups under terminal reward.} The all-fail fraction $(1-p)^G$ of
\S\ref{sec:dense:sparsity} stays high even for larger groups, still $44\%$ at $p{=}0.05$ with
$G{=}16$, and even on synthesized terminal tasks \citet{shen2026seta} find that only about half of
all tasks yield any reward variance under binary reward. The same starvation afflicts actor-critic
methods such as PPO~\citep{schulman2017ppo} in a softer form, since a critic fit to near-constant
returns yields near-zero advantages everywhere; GRPO merely makes it exact.

\textbf{Relation to policy-invariant shaping.} PPRS (Eq.~\eqref{eq:shaped}) is not policy-invariant
shaping in the sense of \citet{ng1999policy}. Classical shaping zeroes the potential of the terminal
state, so the shaping terms sum to the same constant $-\Phi(s_0)=0$ on every trajectory and leave the
objective unchanged. We keep the terminal potential, so the shaped return is terminal success plus
$\lambda\,\Phi(s_T)$ and defines a graded objective; zeroing it would give every all-fail group
identical rewards again and reinstate the starvation of \S\ref{sec:dense:sparsity}, which is why the
term is kept. What is preserved is the ranking of trajectories: a passing trajectory earns $1+\lambda$
and a failing one at most $\lambda$, so a deterministic policy that maximizes one objective maximizes
the other, and no trajectory is rewarded for trading completion against partial progress. What is not
preserved is the ranking of stochastic policies in expectation. Let $\pi_1$ and $\pi_2$ pass with
probabilities $p_1<p_2$ and let $\bar\Phi_k\in[0,1]$ be the mean final potential of their failing
rollouts. Then
\[
\mathbb{E}_{\pi_1}[R]-\mathbb{E}_{\pi_2}[R]
=(p_1-p_2)(1+\lambda)+\lambda\bigl[(1-p_1)\bar\Phi_1-(1-p_2)\bar\Phi_2\bigr]
\;\le\;(p_1-p_2)(1+\lambda)+\lambda(1-p_1),
\]
so the shaped objective can prefer the less successful policy only if
$p_2-p_1<\lambda(1-p_1)/(1+\lambda)\le\lambda/(1+\lambda)$, which is $0.2$ at $\lambda=0.25$,
and only when its failures end close to complete while the other's end near empty. Every result we
report is a terminal pass rate under the strict verifier, so any such distortion would appear directly
in the tables; the sparse arm of Table~\ref{tab:ablation}, trained on the same tasks with the terminal
reward alone, is the control. $\lambda$ was fixed at $0.25$ in every run and not tuned.

\textbf{Example progress contract.} For the warehouse environment of
Figure~\ref{fig:data-pipeline} the nodes are \texttt{readable}, \texttt{schema},
\texttt{dimensions}, \texttt{facts}, \texttt{normalization}, \texttt{relations},
\texttt{aggregates}, \texttt{quality\_report}, and \texttt{summary}; \texttt{relations}, for
instance, requires \texttt{dimensions} and \texttt{facts}. The partial scorer $c_v$ uses set
precision and recall for record-valued outputs and declared tolerances for numeric ones, and never
re-runs the strict verifier. Each node's credit is capped by the weakest of its prerequisites
(\S\ref{sec:dense:potential}), so a correct aggregate built on a wrong fact table earns nothing.

\textbf{Measured groups.} To see the starvation of \S\ref{sec:dense:sparsity} on real rollouts, we
graded the workspace after every turn for $G{=}8$ rollouts on each of 12 seeds of the warehouse
environment, for the untrained Qwen3.8-27B and for the trained 27B checkpoint, under the training
harness (65k-token context, 60 tool calls), which grades the terminal reward only when the policy
issues the submit command. Terminal reward is $0$ on all 192 rollouts, so every one of the 24 groups
has zero advantage under Eq.~\eqref{eq:grpo-adv}; yet under $\Phi$ every group has non-zero
variance, with rollouts ending on the prerequisite-gated plateaus $0.05$ (readable and schema
partially present), $0.26$ (every table partially loaded), $0.69$ (schema and facts complete),
$0.78$, $0.90$, and $1$. Figure~\ref{fig:potential} shows one of these groups.

\textbf{Shaping within a failing group.} In a group in which no rollout passes, $\lambda$ cancels and the
potential \emph{is} the objective; \citet{wang2026darkroom} shows that GRPO agents collapse when
this happens with a shaping term that is not aligned with the task, and it is safe here only
because $\Phi$ is exact progress toward the ground-truth world. Among failing rollouts, the shaped
reward therefore imposes a total order by how much of the workflow they completed, and it penalizes
regressions, such as overwriting a correct table with a corrupted one, through negative increments
$r_t<0$. The fraction of groups that yield
gradient rises from $1-(1-p)^G$ to the fraction whose rollouts are not all stalled at the same
stage. In training, $R(\tau)$ enters with per-turn credit
(Appendix~\ref{app:train}): turn $t$ receives the increment $\lambda r_t$ and the final turn additionally
receives the terminal term.

\textbf{Relation to potential-based shaping.} In the language of potential-based
shaping~\citep{ng1999policy}, $\Phi$ is the potential and $r_t$ its difference. Classical shaping
requires that the shaped objective have the same optimal policies as the original; we deliberately
do not, because the whole point is that the shaped objective is \emph{not} indifferent to partial
progress. What we retain is agreement on maximizers (any passing rollout outscores any failing one)
and anchoring ($\Phi=1$ exactly when the verifier passes), so the acceptance set of a task is
unchanged. This is the opposite route from the usual remedy for tasks beyond a learner's frontier,
which is to simplify, scaffold, or defer the task until some rollout
succeeds~\citep{florensa2018goalgan,chen2025agentfrontier,wolf2026frontier,zeng2026ace}; we leave
the task and the harness fixed and raise the resolution of the reward. We prefer a task-aligned
potential to the test-pass fraction of \citet{shen2026seta} and the passing-assertion count of \citet{yang2026t1}, which score
only what the terminal tests probe and carry no notion of regression. The signal is also denser
\emph{within} a group that contains a success: the lone passing rollout is contrasted not with an
undifferentiated mass of failures but with a graded spectrum of near-misses. Because $\Phi$ is
produced by a scorer with a code closure disjoint from the verifier's and has passed the causality
and exploit-resistance gates of Appendix~\ref{app:data:gates}, none of the probed gaming strategies
applied to the public tree earns $\Phi>0$, and $\Phi$ rises only when the output tree matches more
of the workflow's correct intermediate and final artifacts.

\section{Training and Evaluation Details}
\label{app:train}

\subsection{Training Recipe and Hyperparameters}
\vspace{-2mm}
\label{app:train:recipe}

We optimize with DPPO: group-centered advantages, a total-variation divergence bound of $0.1$ on
the policy ratio, and no KL anchor. Groups whose rewards have zero variance are discarded and refilled by
active sampling, so every step sees 8 informative groups. The reward is Eq.~\eqref{eq:shaped} with $\lambda=0.25$, applied with
per-turn credit: turn $t$ receives the increment $\lambda r_t$, the final turn additionally receives the
terminal term, and each turn's advantage is its return-to-go minus the group mean of that return at
the same turn index. Table~\ref{tab:hparams} lists the recipe shared by every run; runs differ only
in the corpus subset, the reward mechanism, the model size, and the hardware they ran on.

\begin{table}[h]
\centering
\small
\caption{\textbf{Training hyperparameters} shared by every RL run in this paper.}
\vspace{-3mm}
\label{tab:hparams}
\begin{tabular}{l p{0.7\textwidth}}
\toprule
Setting & Value \\
\midrule
Objective & group-relative policy gradient with centered advantages (Eq.~\eqref{eq:grpo-adv}) \\
Trust region & total-variation divergence bound $0.1$ on the policy ratio (DPPO loss); no KL anchor ($\beta{=}0$) \\
Rollouts per task ($G$) & 24 \\
Unique tasks per step & 8 (192 rollouts per step) \\
Active sampling & zero-variance groups discarded and refilled \\
Optimizer & AdamW, learning rate $10^{-6}$, constant schedule, one epoch per batch \\
Sampling temperature & 1.0 \\
Prompt / response budget & 2{,}048 prompt tokens; 65{,}536 response tokens \\
Per-turn budget & 16{,}384 tokens; at most 64 tool calls per rollout \\
Tools & a single \texttt{bash} tool in a per-task container (\texttt{python:3.13-slim} base image) \\
Reward & Eq.~\eqref{eq:shaped} with $\lambda{=}0.25$; per-turn credit, return-to-go with $\gamma{=}1$, per-turn-index group baseline \\
Terminal reward & strict verifier score at submission, $1.0$ for a pass \\
Precision / parallelism & bf16 with fp32 LM head; ZeRO-3, sequence parallelism 4 (9B) \\
Learner / inference split & 16 learner GPUs + 16 inference engines on 32-GPU runs; 16 + 48 on the 64-GPU 27B run \\
Seed & 42 \\
\bottomrule
\end{tabular}
\vspace{-5mm}
\end{table}

\subsection{Evaluation Protocol}
\vspace{-2mm}
\label{app:eval}

\evalsuite consists of 18 expert-authored long-horizon tasks in three capability
categories (Math \& Science, 8 tasks; ML, 5; Coding \& System, 5), each graded in $[0,1]$ by its own
strict grader. Every Qwen model, base or trained, runs in \emph{Vanillux2}, the single-\texttt{bash}
harness used in training, with 64 turns and 16{,}384 tokens per turn and 5 independent rollouts per
task. Frontier models run in their native agent harnesses (Codex, Claude Code, Gemini CLI), the
setting most favorable to them; the largest open model, Qwen3.8-2.4T-A95B, runs in Vanillux2. We
report the mean graded score (\textbf{S@1}), the fraction of rollouts that fully solve the task
(\textbf{P@1}), and their best-of-5 counterparts (\textbf{S@5}, \textbf{P@5}). On
TB~3.0 and~4.0~\citep{merrill2026terminalbench,marten2026tb3,marten2026tb4} we evaluate the ``CM'' subsets (22 and 18
tasks; selection described below); on
TB~2.1 (89 tasks) we use the benchmark's reference agent Terminus-2 for the 27B models
and Vanillux2 for the 9B models, with $k{=}5$. Task selection and scoring conventions are given in the table captions.

\textbf{Selection of the ``CM'' subsets and task lists.}
\label{app:eval:tasks}
Our training generators cover one domain, computing and math, while TB~3.0 spans many, so we evaluate
on the TB tasks that fall in that domain. To select them we gave Claude the 74 TB~3.0 task directories
together with a description of our computing-and-math domains and asked it to judge, task by
task, whether the task belongs to that domain even just slightly; we keep the in-domain tasks that our CPU-only harness can
run and grade. This gives the 22-task TB~3.0 subset, and the 18 of these tasks that remain in TB~4.0
form the TB~4.0 subset. The computing-and-math split of Agents' Last Exam has 23 tasks, and \evalsuite
keeps the 18 that run in a Linux Docker container, the sandbox of our training and evaluation harness.
The other five need a full virtual machine, which that sandbox does not provide. Four boot a
Windows~10 guest to analyze Windows binaries or packet captures with desktop tools such as IDA,
Ghidra, x64dbg, and Wireshark (\texttt{ghidra\_malware\_config\_extraction\_01},
\texttt{pcap\_enterprise\_triage\_01}, \texttt{tris\_crackme}, \texttt{newyear\_keygen2}); the fifth,
\texttt{k8s\_migration\_1}, boots an Ubuntu~22 guest to run a Minikube cluster on Docker, which would
require Docker inside our container. The exclusion follows from the runtime alone, not from task
difficulty, and every model in Table~\ref{tab:main_results}, ours and frontier, is scored on the same
18 tasks. The 18 \evalsuite tasks are listed in Table~\ref{tab:ale-cm-tasks} and the TB
subsets in Table~\ref{tab:tb-cm-tasks}.

\begin{table}[htbp]
\centering
\footnotesize
\setlength{\tabcolsep}{4pt}
\caption{\textbf{The 18 \evalsuite tasks} (computing-and-math split of Agents' Last Exam), grouped by the
capability category used in Table~\ref{tab:main_results}.}
\vspace{-2mm}
\label{tab:ale-cm-tasks}
\resizebox{\textwidth}{!}{%
\begin{tabular}{@{}lll@{}}
\toprule
Category & Task & Title \\
\midrule
Math \& Science (8) & \texttt{branch\_bound\_atsp} & Branch-and-Bound ATSP Solver \\
 & \texttt{cfr\_game\_theory\_equilibrium} & CFR Game Theory Equilibrium \\
 & \texttt{particle\_filter\_nonlinear\_tracking} & Particle Filter Nonlinear Tracking \\
 & \texttt{k3\_abelian\_extensions} & Finite Abelian Extension Classification \\
 & \texttt{synthetic\_causal\_structure\_inference} & Synthetic Causal Structure Inference \\
 & \texttt{clustered\_cyclic\_code\_circuit\_level\_simulation} & Clustered-Cyclic Code Circuit Simulation \\
 & \texttt{ising\_post\_measurement\_1} & Post-Measurement States for a Quantum Ising Chain \\
 & \texttt{go\_game\_reconstruction\_1} & Go Game Reconstruction 1 \\
\addlinespace
ML (5) & \texttt{dit\_pipeline\_cfg\_alignment\_fid\_256\_001} & Repair Standalone DiT Pipeline CFG Alignment \\
 & \texttt{paper\_reproduction\_instance\_1} & LCA-on-the-Line Table 2 Reproduction \\
 & \texttt{ranking\_node\_feature\_parity\_recovery\_instance\_1} & Ranking Node Feature Parity Recovery \\
 & \texttt{mp\_checkpoint\_consolidation\_v2} & Checkpoint Consolidation V2 \\
 & \texttt{recsys\_cold\_start\_instance\_1} & Hybrid Recommender Cold-Start Benchmark \\
\addlinespace
Coding \& System (5) & \texttt{k8s\_payment\_api\_root\_cause\_analysis} & Kubernetes Payment API Root Cause Analysis \\
 & \texttt{os\_log\_permission\_guard\_v1} & OS Log Permission Guard \\
 & \texttt{data\_pipeline\_etl\_instance\_1} & Retail ETL Warehouse Cleanup \\
 & \texttt{cp\_test\_gen\_1} & Competitive Programming Test Generator -- Binary String Copying \\
 & \texttt{cost\_optimization\_1} & AWS Cost Optimization Analysis \\
\bottomrule
\end{tabular}}
\end{table}

\begin{table}[htbp]
\centering
\footnotesize
\setlength{\tabcolsep}{4pt}
\caption{\textbf{The Terminal-Bench ``CM'' subsets} evaluated in Table~\ref{tab:tbench}: the 22 TB~3.0
tasks in the computing-and-math domain, with the official TB category and whether the task remains in
TB~4.0 (18 do).}
\vspace{-2mm}
\label{tab:tb-cm-tasks}
\begin{tabular}{@{}llc@{}}
\toprule
TB task & TB category & In TB~4.0 \\
\midrule
\texttt{batched-eval-parity} & ML / Evaluation & yes \\
\texttt{cargo-flight-dispatch} & Operations / Logistics & yes \\
\texttt{cli-2ph-simplex} & Software / Algorithms & -- \\
\texttt{ctr-optimization} & Operations / Marketing & yes \\
\texttt{data-anonymization} & Software / Data engineering & yes \\
\texttt{distributed-dedup} & Software / Systems & yes \\
\texttt{fix-uautomizer-soundness} & Software / Languages & -- \\
\texttt{freight-dispatch-shift} & Operations / Logistics & yes \\
\texttt{ico-path-patch} & Security / Reverse engineering & -- \\
\texttt{live-database-cutover} & Software / Databases & yes \\
\texttt{memcached-backdoor} & Security / Reverse engineering & -- \\
\texttt{mp-checkpoint-consolidation} & ML / Inference & yes \\
\texttt{mvcc-lsm-compaction} & Software / Databases & yes \\
\texttt{payments-pipeline-fix} & Software / Systems & yes \\
\texttt{pretrain-shard-corruption} & ML / Training & yes \\
\texttt{production-planning} & Operations / Supply chain & yes \\
\texttt{risk-scorer-replay} & ML / Evaluation & yes \\
\texttt{session-window-debug} & Software / Systems & yes \\
\texttt{shadow-relay} & Security / Forensics & yes \\
\texttt{telecom-entity-resolution} & Software / Data engineering & yes \\
\texttt{uefi-bootkit} & Security / Forensics & yes \\
\texttt{wal-recovery-ordering} & Software / Databases & yes \\
\bottomrule
\end{tabular}
\end{table}

\textbf{Representativeness of the ``CM'' subsets.}
\label{app:eval:subset}
Tables~\ref{tab:tb3-subset-vs-full} and~\ref{tab:tb4-subset-vs-full} compare, for every official
leaderboard row of Table~\ref{tab:tbench}, its score on our 22-task (TB~3.0) and 18-task
(TB~4.0) subsets with its score on the full 74- and 66-task benchmarks. Both sides are
computed from the same official per-trial results (5 trials per task, ungraded trials scored 0,
P@1 the per-task mean reward and P@5 the unbiased pass@5 estimator, macro-averaged over tasks); the
full-set P@1 reproduces the leaderboard's published accuracy to within $0.01$ (the residual comes
from trials the leaderboard re-scored after our snapshot). GLM~5.2 has no TB~4.0 entry,
and our own Gemini and Qwen runs are not leaderboard rows, so they are omitted. Subset scores run
somewhat above full-set scores for most models, but the two agree closely on how models compare:
across models, Pearson $r$ (linear agreement of the scores) is $0.96$ (P@1) and $0.89$ (P@5) on
TB~3.0 and $0.97$ and $0.92$ on 4.0, and Spearman $\rho$ (agreement of the model
ranking) is $0.94$/$0.83$ and $0.90$/$0.91$.

\begin{table}[htbp]
\centering
\footnotesize
\setlength{\tabcolsep}{3pt}
\begin{minipage}[t]{0.49\textwidth}
\centering
\captionof{table}{\textbf{TB 3.0: our 22-task ``CM'' subset versus the full 74-task
benchmark}, official leaderboard rows of Table~\ref{tab:tbench}. ``Official acc.'' is the published
leaderboard accuracy. Across the 10 models: Pearson $r{=}0.96$ (P@1) and $0.89$ (P@5); Spearman
$\rho{=}0.94$ and $0.83$.}
\vspace{-3mm}
\label{tab:tb3-subset-vs-full}
\begin{tabular}{l cc cc c}
\toprule
\multirow{2}{*}{Model} & \multicolumn{2}{c}{Subset (22)} & \multicolumn{2}{c}{Full (74)} & Official \\
\cmidrule(lr){2-3}\cmidrule(lr){4-5}
 & P@1 & P@5 & P@1 & P@5 & acc. \\
\midrule
Claude Fable 5 & 0.41 & 0.64 & 0.34 & 0.61 & 0.34 \\
Claude Opus 5 & 0.53 & 0.64 & 0.42 & 0.58 & 0.43 \\
GPT-5.6 (sol) & 0.41 & 0.55 & 0.34 & 0.55 & 0.35 \\
GPT-5.6 (terra) & 0.30 & 0.55 & 0.20 & 0.38 & 0.21 \\
GPT-5.6 (luna) & 0.18 & 0.32 & 0.13 & 0.27 & 0.14 \\
Grok 4.6 & 0.30 & 0.50 & 0.26 & 0.45 & 0.26 \\
Claude Sonnet 5 & 0.16 & 0.32 & 0.14 & 0.34 & 0.15 \\
Claude Opus 4.8 & 0.19 & 0.50 & 0.21 & 0.42 & 0.21 \\
Grok 4.5 & 0.26 & 0.55 & 0.15 & 0.34 & 0.16 \\
GLM 5.2 & 0.04 & 0.18 & 0.04 & 0.14 & 0.05 \\
\bottomrule
\end{tabular}
\end{minipage}\hfill
\begin{minipage}[t]{0.49\textwidth}
\centering
\captionof{table}{\textbf{TB 4.0: our 18-task ``CM'' subset versus the full 66-task
benchmark}, same rows and protocol as Table~\ref{tab:tb3-subset-vs-full} (GLM~5.2 has no 4.0
entry). Across the 9 models: Pearson $r{=}0.97$ (P@1) and $0.92$ (P@5); Spearman $\rho{=}0.90$ and
$0.91$.}
\vspace{-3mm}
\label{tab:tb4-subset-vs-full}
\begin{tabular}{l cc cc c}
\toprule
\multirow{2}{*}{Model} & \multicolumn{2}{c}{Subset (18)} & \multicolumn{2}{c}{Full (66)} & Official \\
\cmidrule(lr){2-3}\cmidrule(lr){4-5}
 & P@1 & P@5 & P@1 & P@5 & acc. \\
\midrule
Claude Fable 5 & 0.46 & 0.61 & 0.45 & 0.68 & 0.45 \\
Claude Opus 5 & 0.52 & 0.72 & 0.52 & 0.71 & 0.52 \\
GPT-5.6 (sol) & 0.39 & 0.61 & 0.37 & 0.61 & 0.37 \\
GPT-5.6 (terra) & 0.23 & 0.44 & 0.22 & 0.44 & 0.22 \\
GPT-5.6 (luna) & 0.18 & 0.33 & 0.17 & 0.33 & 0.17 \\
Grok 4.6 & 0.31 & 0.50 & 0.20 & 0.39 & 0.20 \\
Claude Sonnet 5 & 0.19 & 0.50 & 0.12 & 0.35 & 0.12 \\
Claude Opus 4.8 & 0.27 & 0.50 & 0.24 & 0.47 & 0.24 \\
Grok 4.5 & 0.20 & 0.33 & 0.12 & 0.27 & 0.12 \\
\bottomrule
\end{tabular}
\end{minipage}
\vspace{-1mm}
\end{table}

For \evalsuite the official Agents' Last Exam leaderboard publishes only aggregate scores per model
and harness (mean score and pass rate over the full 152-task benchmark, one run per task) and no
per-task results, so Table~\ref{tab:ale-subset-vs-full} sets our own Table~\ref{tab:main_results}
runs on the 18-task split against the leaderboard's headline entry for the same model in the same
harness family (its default row, the reasoning-effort variant with the highest pass rate, retrieved
2026-09-18). Five frontier rows of Table~\ref{tab:main_results} have such an entry; Claude Sonnet~5,
the Gemini Flash models, and Qwen3.8-2.4T-A95B do not. Because the two sides differ in harness
configuration and number of rollouts, agreement is looser than on TB: across the five
models Spearman $\rho$ is $0.95$ for pass rate and $0.63$ for score (Pearson $r$ $0.75$ and $0.56$),
with Gemini~3.1~Pro the main outlier, scoring far higher in our Gemini CLI runs on the split than on
the full benchmark. Close agreement is neither expected nor our aim: the full benchmark spans many
domains beyond computing and math, such as engineering, finance, multimodal, and 3D tasks, whereas
\evalsuite isolates the one domain our data targets.

\begin{table}[H]
\centering
\setlength{\tabcolsep}{5pt}
\small
\caption{\textbf{Agents' Last Exam: our 18-task \evalsuite split (Table~\ref{tab:main_results}, 5
rollouts per task) versus the official full 152-task leaderboard} (one run per task; the leaderboard's
default row for the model in the same harness family, with its reasoning-effort setting). Across the 5
models: Pearson $r{=}0.56$ (score) and $0.75$ (pass rate); Spearman $\rho{=}0.63$ and $0.95$.}
\vspace{-3mm}
\label{tab:ale-subset-vs-full}
\begin{tabular}{l l cc cc l}
\toprule
\multirow{2}{*}{Model} & \multirow{2}{*}{Harness} & \multicolumn{2}{c}{Our \evalsuite split (18 tasks)} & \multicolumn{2}{c}{Official full ALE (152 tasks)} & Official \\
\cmidrule(lr){3-4}\cmidrule(lr){5-6}
 & & S@1 & P@1 & Score & Pass rate & effort \\
\midrule
GPT-5.6 (sol) & Codex & 0.58 & 0.28 & 0.54 & 0.31 & xhigh \\
GPT-5.6 (terra) & Codex & 0.50 & 0.22 & 0.51 & 0.28 & max \\
Claude Opus 5 & Claude Code & 0.58 & 0.28 & 0.55 & 0.32 & high \\
Claude Opus 4.8 & Claude Code & 0.52 & 0.22 & 0.45 & 0.27 & max \\
Gemini 3.1 Pro & Gemini CLI & 0.52 & 0.21 & 0.33 & 0.16 & default \\
\bottomrule
\end{tabular}
\vspace{-5mm}
\end{table}

\subsection{Qualitative Examples on \evalsuite}
\vspace{-2mm}
\label{app:eval:qualitative}

Table~\ref{tab:main_results} aggregates over 18 tasks and 5 rollouts; here we trace single rollouts on three more tasks in the layout of
Figure~\ref{fig:qual-example}, which shows the \texttt{k8s\_payment\_api\_root\_cause\_analysis} task: a rollout of the strongest frontier agent (Claude Fable~5), a rollout of a second frontier agent
(GPT-6 Astra or GPT-5.6), a typical rollout of the untrained Qwen3.8-27B base model, and a rollout of \model-27B.
The tasks are those on which our rollout beats the base rollout; Table~\ref{tab:qual-scoreboard} lists the graded scores. Every Qwen
rollout ran in Vanillux2, a single \texttt{bash} tool with 64 turns; frontier rollouts ran in their native harnesses. Every task
below is also solved by at least one frontier agent, so these are not tasks on which our model is uniquely capable; the figures
show \emph{how} the rollouts differ. Green marks a decision that earns grader credit, red one that forfeits it, grey lines
summarize routine steps, and \texttt{t} is the trajectory step.

\begin{table}[H]
\centering
\footnotesize
\setlength{\tabcolsep}{4pt}
\caption{\textbf{Graded scores of the rollouts traced in Figure~\ref{fig:qual-example} and Figures~\ref{fig:app-qual-ising}--\ref{fig:app-qual-k3}.} Each cell is the
score of the single rollout shown in that column; frontier rollouts in their native harness, Qwen rollouts in Vanillux2.}
\vspace{-2mm}
\label{tab:qual-scoreboard}
\resizebox{\textwidth}{!}{%
\begin{tabular}{l l l l c c}
\toprule
Task & Category & Best frontier rollout & Second frontier rollout & Base rollout & Ours \\
\midrule
\texttt{k8s\_payment\_api\_root\_cause\_analysis} & Coding \& Sys. & 0.97 (Claude Fable 5) & 0.75 (GPT-6 Astra) & 0.70 & \textbf{0.97} \\
\texttt{ising\_post\_measurement\_1} & Math \& Sci. & 1.00 (Claude Fable 5) & 1.00 (GPT-5.6 sol) & 0.00 & \textbf{1.00} \\
\texttt{branch\_bound\_atsp} & Math \& Sci. & 1.00 (Claude Fable 5) & 1.00 (GPT-5.6 sol) & 0.00 & \textbf{1.00} \\
\texttt{k3\_abelian\_extensions} & Math \& Sci. & 1.00 (Claude Fable 5) & 1.00 (GPT-5.6 sol) & 0.00 & \textbf{1.00} \\
\bottomrule
\end{tabular}%
}
\vspace{-3mm}
\end{table}

\begin{figure}[htbp]
\centering
\scriptsize
\setlength{\fboxsep}{2pt}
\begin{minipage}[t]{0.238\textwidth}
\qhead{Claude Fable 5}{1.00}
\qstep{11--19}{No ancilla file is staged $\to$ critical-ancilla rule; installs the pinned NumPy/SciPy with \texttt{uv}.}
\qkey{columngreen}{20--33}{Trotterized imaginary-time evolution ($\Delta t{=}0.0002$, $K{=}50000$) for the ground state, cross-checked against sparse exact diagonalization: energy $-12.78491$ both ways, infidelity $1.0{\times}10^{-7}$; computes $a(j)$, $a(j,k)$, $V_{jk}$, $m_j$, $H'$, $H_m$ for all 1{,}024 outcomes by bit-flip indexing.}
\qstep{33--36}{Verifies shapes and dtypes, unit trace and Hermiticity of every $\rho_1$, correlators recomputed from $\rho_1$; submits.}
\end{minipage}\hfill
\begin{minipage}[t]{0.238\textwidth}
\qhead{GPT-5.6 (sol)}{1.00}
\qstep{3--9}{Reads specification and config; the ancilla is unstaged; installs the pinned dependencies.}
\qkey{columngreen}{20--21}{Exact diagonalization (\texttt{eigsh}) of the spec Hamiltonian; builds $H'$ and $H_m$ with the big-endian conventions; checks normalization, Hermiticity and correlator consistency.}
\qstep{23--26}{Independent reload check: shapes, dtypes, positivity, Hamiltonian residual $8.1{\times}10^{-13}$, probability sum $1-2{\times}10^{-16}$; submits.}
\end{minipage}\hfill
\begin{minipage}[t]{0.238\textwidth}
\qhead{Qwen3.8-27B base}{0.00}
\qstep{2--12}{Reads the specification; installs NumPy/SciPy into a venv.}
\qkey{columnred}{14}{\texttt{stage1.py} builds the site operators with \texttt{(idx >{}> b) \& 1} where \texttt{b} is a bit \emph{mask} (512, 256, \dots), not a bit position, so five of the ten $Z_j$ collapse to constants. Its exact diagonalization and its imaginary-time evolution share this table: ``infid ED vs XZ: 3.49e-08'', $E_0=-17.03$ where the true value is $-12.78$.}
\qstep{18--32}{Writes the protocol; fixes shape bugs caught by $|\langle X\rangle|\le 1$ checks.}
\qkey{columnred}{34--42}{``Independent loop-based verification \dots\ matches to machine precision \dots\ E0 from saved psi\_c: -17.0267 (correct ground state).'' The check reuses the same operator table, so agreement proves nothing; submits a state with infidelity $0.79$ to the reference.}
\end{minipage}\hfill
\begin{minipage}[t]{0.238\textwidth}
\qhead{\model-27B (ours)}{1.00}
\qstep{2--14}{Reads specification and config; installs the pinned NumPy/SciPy into a \texttt{uv} venv.}
\qstep{16--42}{Implements the Trotterized imaginary-time evolution and the vectorized protocol over all 1{,}024 outcomes; fixes a series of tensor-shape bugs.}
\qkey{columngreen}{44--50}{A sanity print flags ``rho1 hermiticity max err: 0.183''; traces it to its own check ($\rho_1^{\mathsf T}$ instead of $\rho_1^{\dagger}$) and confirms $\langle Z\rangle,\langle X\rangle$ read from $\rho_1$ against independently computed correlators to $10^{-15}$.}
\qkey{columngreen}{54--58}{Writes a separate \texttt{crosscheck.py} that rebuilds $H$ from Kronecker products, takes the ground state with \texttt{eigsh}, re-derives every quantity and compares the \emph{saved files}: $p$ $1.1{\times}10^{-4}$, $\rho_1$ $2.9{\times}10^{-4}$, $\langle Z\rangle$ $5.7{\times}10^{-4}$, $\langle X\rangle$ $3.8{\times}10^{-4}$ (tolerances $10^{-4}$ / $10^{-2}$).}
\qstep{60--62}{Confirms output files and untouched inputs; submits.}
\end{minipage}
\vspace{-1mm}
\caption{\textbf{Qualitative example on the \evalsuite task \texttt{ising\_post\_measurement\_1}} (Math \& Science). Compute the exact
ground state of a 10-qubit critical transverse-field Ising chain, couple it to an ancilla copy and derive, for all $2^{10}$
projective ancilla outcomes, the post-measurement probabilities, site-1 reduced density matrices and one-body correlators; four
arrays are graded against a hidden reference with binary credit. \colorbox{columngreen}{Green} marks a decision that secures the
result, \colorbox{columnred}{red} one that forfeits it. All three solving rollouts cross-check the ground state with an
\emph{independent} construction of $H$; the base model's exact-diagonalization and imaginary-time codes share one buggy operator
table, so their agreement proves nothing.}
\label{fig:app-qual-ising}
\end{figure}

\begin{figure}[htbp]
\centering
\scriptsize
\setlength{\fboxsep}{2pt}
\begin{minipage}[t]{0.238\textwidth}
\qhead{Claude Fable 5}{1.00}
\qstep{13--25}{Sets up the venv; writes an assignment-relaxation branch-and-bound with Carpaneto--Toth subtour branching; fixes a stray non-ASCII identifier.}
\qstep{29--35}{Tiers 1 and 2 proven optimal in 29 and 657 nodes; tier 3 runs in the background.}
\qkey{columngreen}{39--47}{Tier 3 lands at 0.49999\,\%, ``just under the 0.005 threshold'' $\to$ independently recomputes every tour and cost, then confirms tier 1 against an exact Held--Karp dynamic program.}
\qstep{51}{Writes \texttt{results.json} (tier 3: 172{,}168 nodes, 100\,s); submits.}
\end{minipage}\hfill
\begin{minipage}[t]{0.238\textwidth}
\qhead{GPT-5.6 (sol)}{1.00}
\qstep{6--16}{Subtour-branching solver with Hungarian bounds and directed local search; tier 1 proven optimal in 27 nodes.}
\qkey{columngreen}{19--31}{``The first branching scheme is too redundant on Tier 2'' $\to$ finds the bug (already-forced arcs re-branched), makes the partition finite; a Lagrangian bound closes tier 3 immediately.}
\qstep{36--42}{Validates every tour and gap by independent recomputation (tier 3 gap 0.2498\,\%); cleans the temporary runtime; submits.}
\end{minipage}\hfill
\begin{minipage}[t]{0.238\textwidth}
\qhead{Qwen3.8-27B base}{0.00}
\qstep{8--18}{Finds SciPy in the system Python; nearest-neighbor $+$ 2-opt incumbents; benchmarks the Hungarian solver.}
\qkey{columnred}{24--32}{Assignment-bound branch-and-bound with shape, heap-sign and int8-overflow bugs; the pure assignment bound (330 vs.\ optimum 411.7) leaves a 20\,\% gap and the 12-city search explodes to 435k nodes.}
\qstep{40--58}{Held--Karp confirms the 12-city optimum; the Lagrangian bound is rewritten four times (dual sign, multipliers stuck at 0, overflow) until at t58 it proves tier 1 at the root and bounds tier 3 at 879.98.}
\qkey{columnred}{60--64}{Turns to incumbent heuristics (subtour patching, Or-opt) instead of writing the report; the context window overflows with no \texttt{results.json}.}
\end{minipage}\hfill
\begin{minipage}[t]{0.238\textwidth}
\qhead{\model-27B (ours)}{1.00}
\qstep{2--22}{Sets up the venv; generates the instances; verifies genuine asymmetry.}
\qstep{28--36}{Best-first branch-and-bound with assignment relaxation: 12 cities in 13 nodes, 20 cities in 378 nodes; notices the reported bound lags the optimum and fixes the bound bookkeeping.}
\qkey{columngreen}{42--58}{Held--Karp DP confirms the 12-city optimum; the plain bound stalls at 1.95\,\% on 35 cities after 360k nodes $\to$ a Lagrangian subgradient bound closes tier 3 to 0.0\,\% in 4 nodes.}
\qkey{columngreen}{72--84}{Writes \texttt{results.json} first, then verifies: independent recomputation of every tour and cost, Held--Karp cross-check for tier 1, an independent heuristic for tier 3, and a determinism re-run (diff only in \texttt{runtime\_seconds}).}
\qstep{86}{Submits.}
\end{minipage}
\vspace{-1mm}
\caption{\textbf{Qualitative example on the \evalsuite task \texttt{branch\_bound\_atsp}} (Math \& Science). Implement a
branch-and-bound solver in Python/NumPy/SciPy for three asymmetric TSP instances: 12 and 20 cities to proven optimality, 35 cities to
within a 0.5\,\% gap, reported in a JSON file checked against hidden optima. \colorbox{columngreen}{Green} marks a decision that
secures the result, \colorbox{columnred}{red} one that forfeits it. The three solving rollouts write the deliverable and then verify
it independently; the base model has a working bound by t58 but keeps improving heuristics and never writes the file.}
\label{fig:app-qual-bb}
\end{figure}

\begin{figure}[htbp]
\centering
\scriptsize
\setlength{\fboxsep}{2pt}
\begin{minipage}[t]{0.238\textwidth}
\qhead{Claude Fable 5}{1.00}
\qstep{2--30}{Reads the specification; GAP enumeration: for each $m$, all abelian groups of order $64m$ by elementary divisors, exhaustive check for a cyclic subgroup $N$ of order $m$ with quotient $\cong(\mathbb Z/4)^3$ $\to$ 14 extensions.}
\qkey{columngreen}{30--31}{Product-type classification in Python via Krull--Schmidt: every partition of the elementary-divisor multiset is checked against the five allowed forms $\to$ 2 non-product extensions ($[4,4,4,4]$, $[4,4,4,8]$).}
\qstep{35}{Verifies the output file; submits.}
\end{minipage}\hfill
\begin{minipage}[t]{0.238\textwidth}
\qhead{GPT-5.6 (sol)}{1.00}
\qstep{6--13}{GAP enumeration that tests cyclic kernels directly $\to$ 14 extension groups.}
\qkey{columngreen}{13--16}{Exhaustive product-family check ``with special attention to the two rank-four cases that may fail'' $\to$ exactly two non-product extensions.}
\qstep{16--22}{Validates schema, canonical invariant factors, orders and sorting; submits.}
\end{minipage}\hfill
\begin{minipage}[t]{0.238\textwidth}
\qhead{Qwen3.8-27B base}{0.00}
\qstep{2--40}{Reads the specification; GAP scripts fail on syntax (\texttt{Partitions}, \texttt{min}, list constructors) and are fixed one at a time.}
\qstep{44--118}{The enumeration works by t118: ``m=8: 3 extensions \dots\ [4, 4, 4, 8] order=512 pt=false \dots''}
\qkey{columnred}{120--128}{Starts a brute-force verifier \texttt{verify\_all.gap} for the product-type rule and debugs its syntax instead of writing the result.}
\qkey{columnred}{130}{``max\_turns (64) reached without submitting'': no \texttt{results.json} is ever written.}
\end{minipage}\hfill
\begin{minipage}[t]{0.238\textwidth}
\qhead{\model-27B (ours)}{1.00}
\qstep{2--12}{Reads specification and config; GAP 4.11 is present.}
\qstep{14--46}{The first GAP script fails on missing built-ins (\texttt{IsIsomorphic}, \texttt{QuotientGroup}, \texttt{Partitions}); probes what the installation provides.}
\qkey{columngreen}{48--68}{Builds its own isomorphism test for abelian 2-groups from elementary-divisor multisets; a debug script isolates an \texttt{Htype} indexing bug.}
\qkey{columngreen}{84--100}{Rewrites the invariant-factor enumeration and the product-type classification; \texttt{results.json}: 14 extensions, 2 non-product.}
\qstep{120--130}{Validates keys, ordering and canonical form; submits.}
\end{minipage}
\vspace{-1mm}
\caption{\textbf{Qualitative example on the \evalsuite task \texttt{k3\_abelian\_extensions}} (Math \& Science). Enumerate every
finite abelian $G$ in $0\to\mathbb Z/m\to G\to(\mathbb Z/4)^3\to 0$ for $m=1,\dots,8$ in GAP, classify product type, and write an
exact-match JSON. \colorbox{columngreen}{Green} marks a decision that secures the result, \colorbox{columnred}{red} one that
forfeits it. The GAP build lacks \texttt{IsIsomorphic}; our model works around it and writes the deliverable, whereas the base model
has the enumeration right by t118 but spends its remaining turns on a brute-force verifier and never writes the file.}
\label{fig:app-qual-k3}
\end{figure}

\section{Extended Related Work}
\label{app:related}

\textbf{Synthetic corpora for coding and terminal agents.} The first wave of agent training data
mined real repositories: SWE-bench~\citep{jimenez2024swebench} and SWE-Gym~\citep{pan2024swegym}
collect GitHub issues with their tests, and SWE-rebench~\citep{badertdinov2025swerebench} and
Skywork-SWE~\citep{zeng2025skyworkswe} automate the mining. The second wave synthesizes tasks:
SWE-smith~\citep{yang2025swesmith} injects bugs into working repositories,
SWE-Flow~\citep{zhang2025sweflow} strips a repository along its runtime dependency graph, and
R2E-Gym~\citep{jain2025r2egym} back-translates problem statements from commits. Terminal corpora
follow the same pattern at larger scale, either with a language model writing tasks, containers,
and tests that are then filtered for solvability, as in Endless
Terminals~\citep{gandhi2026endless}, SETA~\citep{shen2026seta}, and recursive solution
extension~\citep{li2026rst} (TMAX~\citep{ivison2026tmax} skips this filter and lets RL drop unsolved tasks), or by reconstructing environments from agent trajectories, as in
CLI-Gym~\citep{lin2026cligym} and Terminal-Universe~\citep{wu2026terminaluniverse}. All of these
scale the number of \emph{instances}: each instance is pull-request-sized or a single session, its
verifier is a test suite that caps the complexity a task can have, and generation depends on a
model whose output must be validated after the fact. We scale \emph{environments}: a coding agent writes each generator once, the authors review
it, and instantiating it involves no model; verification is against the latent world rather than
a test suite. Closer in
spirit is the programmatic synthesis of executable, stateful tool environments,
AutoForge~\citep{cai2025autoforge}, EnvScaler~\citep{song2026envscaler},
ScaleEnv~\citep{tu2026scaleenv}, Agent-World~\citep{dong2026agentworld},
EnvFactory~\citep{xu2026envfactory}, and LOGIGEN~\citep{zeng2026logigen}, which also treats the
environment as the unit of scale~\citep{fang2025agentscaler}; these are API worlds for short
tool-use episodes, and only LOGIGEN emits a progress signal, a per-turn reward from the database's proximity to its target state. Two findings from that line carry over: more environments alone do not help
without difficulty control~\citep{zhu2026beyondscaling}, and coverage rather than count is what
drives transfer~\citep{chen2026dive,chen2026beyondquantity,xu2026envfactory}. Our environments
differ from this family on the two properties it largely lacks: each spans a long, multi-stage workflow
across math and science, ML, and coding and systems domains rather than a short tool-use episode,
and every one ships the progress potential $\Phi_W$ of \S\ref{sec:dense} rather than a bare success
condition, so the difficulty control and dense signal these two findings call for come from the
same generator that supplies coverage.

\textbf{Backward construction.} Latent-world-first synthesis belongs to a long line of methods that
derive a task from its solution. Denoising objectives corrupt text and learn to restore
it~\citep{lewis2019bart}; the data-cleaning community injects errors of controlled repairability
into clean databases to evaluate repair algorithms~\citep{arocena2015bart} and corrupts synthetic
person records to benchmark record linkage~\citep{tran2013geco}; mutation and bug injection corrupt
working code~\citep{yang2025swesmith,hariharan2025breakpoint}; CLI-Gym rewinds a healthy
environment to an earlier broken state~\citep{lin2026cligym}; instruction back-translation derives
prompts from answers~\citep{li2023backtranslation,wu2024inversecoder}; and procedural generators
with rule-based verifiers supply unlimited single-turn reasoning problems whose answers are known by
construction~\citep{stojanovski2025reasoninggym,chen2025enigmata,liu2025synlogic,li2025internbootcamp}.
In every case the corrupted object is a single artifact graded by a pre-existing oracle, or a
one-step puzzle. We corrupt a rendering of an entire world into many mutually constraining files of
different formats, and the oracle is the world itself; this is what lets exactness, and the ground
truth of every intermediate stage, survive a multi-hour horizon. In the taxonomy of
\citet{zeng2026ace}, generation pipelines anchor on the environment, the task, the trajectory, or a
blueprint that coordinates them; anchoring on the verified terminal state itself is what removes
the artifact drift among task, environment, and verifier that separately generated factors
invite~\citep{ivanov2026anchor}. Concretely, we anchor one level earlier than any object in that
taxonomy: the latent world $W$ itself, so that the task $x$, the input tree $s_0$, the strict
verifier $\operatorname{ver}_W$, and the progress potential $\Phi_W$ (Eq.~\eqref{eq:env}) are four
views of one object rather than four artifacts a blueprint must keep synchronized after the fact.
CLI-Gym is the closest of these in mechanism, but it rewinds a single running environment to an
earlier state and stops at one pass/fail check, with no signal over the stages in between.

\textbf{World-state verification, environment scaling, and long-horizon benchmarks.} Verifying an
agent against the final state of a simulated world rather than against its transcript is
established in AppWorld~\citep{trivedi2024appworld},
$\tau$-bench~\citep{yao2024taubench}, WebArena~\citep{zhou2023webarena}, and
OSWorld~\citep{xie2024osworld}, all with hand-authored tasks for evaluation. Recent work generates
the world too: AgentMercury~\citep{jeong2026agentmercury} has a model author persistent business
worlds with executable invariants, and \citet{zhang2026trustworthyworlds} derive dense rewards from
progress predicates over generated web sites; neither uses a declared corruption, and both depend
on a model to write the world. A growing literature argues that environments are the bottleneck of
agentic RL~\citep{xi2025agentgymrl,fang2025agentscaler,song2026envscaler,zhang2025autoenv,gandhi2026endless},
with the caveat that more environments alone do not help without difficulty
control~\citep{zhu2026beyondscaling}. Our thesis is narrower: the environments worth scaling are
those whose latent world supplies both the oracle and the progress signal. Long-horizon benchmarks
with partial credit, TheAgentCompany~\citep{xu2024theagentcompany}, Long-Horizon
Terminal-Bench~\citep{li2026lhterminalbench}, PaperBench~\citep{starace2025paperbench},
MLE-bench~\citep{chan2024mlebench}, and ALE-Bench~\citep{imajuku2025alebench}, target the regime we
train for, but their checkpoints are written per task by hand or graded by a model, and at tens of
instances they are evaluation sets.

\textbf{Sparse reward and dense signal in agentic RL.} Group-relative methods discard rollout
groups with identical rewards~\citep{yu2025dapo,he2025skyworkor1,ivison2026tmax}, schedule prompts
by difficulty~\citep{yang2025kimidev,sha2026thinkprior}, or seed groups with expert
prefixes~\citep{zhang2025bread}; all assume enough solvable prompts to fill a batch. A
complementary line moves the tasks rather than the reward, generating or selecting them near the
learner's current success
band~\citep{florensa2018goalgan,guo2025genenv,chen2025agentfrontier,wolf2026frontier}; this keeps
groups informative but caps difficulty at what the current policy can sometimes finish. Denser
signals come from learned or judged step
rewards~\citep{cui2025prime,choudhury2025agentprm,wang2025sparl,zhou2025sweetrl,zhou2024archer,gunjal2025rubrics},
from advantages over recurring states~\citep{feng2025gigpo}, or from exact but proxy quantities:
similarity to an oracle patch~\citep{wei2025swerl}, the fraction of tests
passed~\citep{du2025swedev,shen2026seta}, the count of passing assertions~\citep{yang2026t1},
compiler diagnostics~\citep{zhang2025rlcsf}, path-centric credit for evidence
gathering~\citep{xia2026searchp1}, or symbolic oracles in closed
domains~\citep{yuan2026vpr}. Test-fraction rewards are the closest to ours and are what lets SETA
and T1 train on terminal tasks, but they score only what the terminal tests probe, are not monotone
in workflow progress, and carry no notion of regression. Our potential is exact progress against the
latent world at every workflow stage, anchored so that $\Phi=1$ exactly when the verifier passes,
and free for every seed. A recent finding shapes how such a signal should be used: a shaping term that is not aligned
with the task collapses GRPO agents, because in all-fail groups it becomes the whole
objective~\citep{wang2026darkroom}. Unlike the learned or judged step rewards above, $\Phi$ needs
no trained critic or LM judge; its relation to classical potential-based
shaping~\citep{ng1999policy} is discussed in Appendix~\ref{app:dense}.

\end{document}